\documentclass{article}

\PassOptionsToPackage{numbers,compress}{natbib}
\usepackage[main,final]{neurips_2026}

\setcitestyle{numbers}
\usepackage{dsfont}
\usepackage{adjustbox}
\usepackage{wrapfig}
\usepackage{booktabs}

\usepackage[utf8]{inputenc} %
\usepackage[T1]{fontenc}    %
\usepackage{hyperref}       %
\usepackage{url}            %
\usepackage{booktabs}       %
\usepackage{amsfonts}       %
\usepackage{nicefrac}       %
\usepackage{microtype}      %
\usepackage{xcolor}         %
\usepackage{graphicx}
\usepackage{amsmath}
\usepackage{algorithm}
\usepackage{algpseudocode}
\usepackage{listings}
\usepackage{wrapfig}
\usepackage{xcolor}
\usepackage{titletoc}
\usepackage{minitoc}
\usepackage{multirow}
\usepackage{array}
\usepackage[table]{xcolor}
\usepackage{arydshln} %
\usepackage{tcolorbox}
\tcbuselibrary{skins, breakable}
\usepackage{listings}

\lstdefinelanguage{json}{
    basicstyle=\ttfamily\footnotesize,
    numbers=none,
    numberstyle=\tiny\color{gray},
    stepnumber=1,
    numbersep=5pt,
    showstringspaces=false,
    breaklines=true,
    frame=single,
    backgroundcolor=\color{gray!10},
    literate=
     *{0}{{{\color{blue}0}}}{1}
      {1}{{{\color{blue}1}}}{1}
      {2}{{{\color{blue}2}}}{1}
      {3}{{{\color{blue}3}}}{1}
      {4}{{{\color{blue}4}}}{1}
      {5}{{{\color{blue}5}}}{1}
      {6}{{{\color{blue}6}}}{1}
      {7}{{{\color{blue}7}}}{1}
      {8}{{{\color{blue}8}}}{1}
      {9}{{{\color{blue}9}}}{1}
      {:}{{{\color{black}:}}}{1}
      {,}{{{\color{black},}}}{1}
      {"}{{{\color{red}"}}}{1},
}

\title{HiPhy: \underline{Hi}erarchical Alignment for \underline{Phy}sically-Plausible Multi-Principle Video Generation}
\author{
\textbf{Tahira Kazimi}\textsuperscript{1} \:
\textbf{Shubhankar Borse}\textsuperscript{2} \:
\textbf{Munawar Hayat}\textsuperscript{2} \:
\textbf{Fatih Porikli}\textsuperscript{2} \:
\textbf{Pinar Yanardag}\textsuperscript{1} \\[0.35em]
\textsuperscript{1}Virginia Tech
\qquad
\textsuperscript{2}Qualcomm AI Research \\[0.7em]
Project Page:
\href{https://hiphy-video.github.io/}
{\texttt{https://hiphy-video.github.io}}
}
\begin{document}

\maketitle
\begin{figure}[h]
\vspace{-1.5em}
    \centering
    \begin{adjustbox}{center}
    \includegraphics[width=\linewidth]{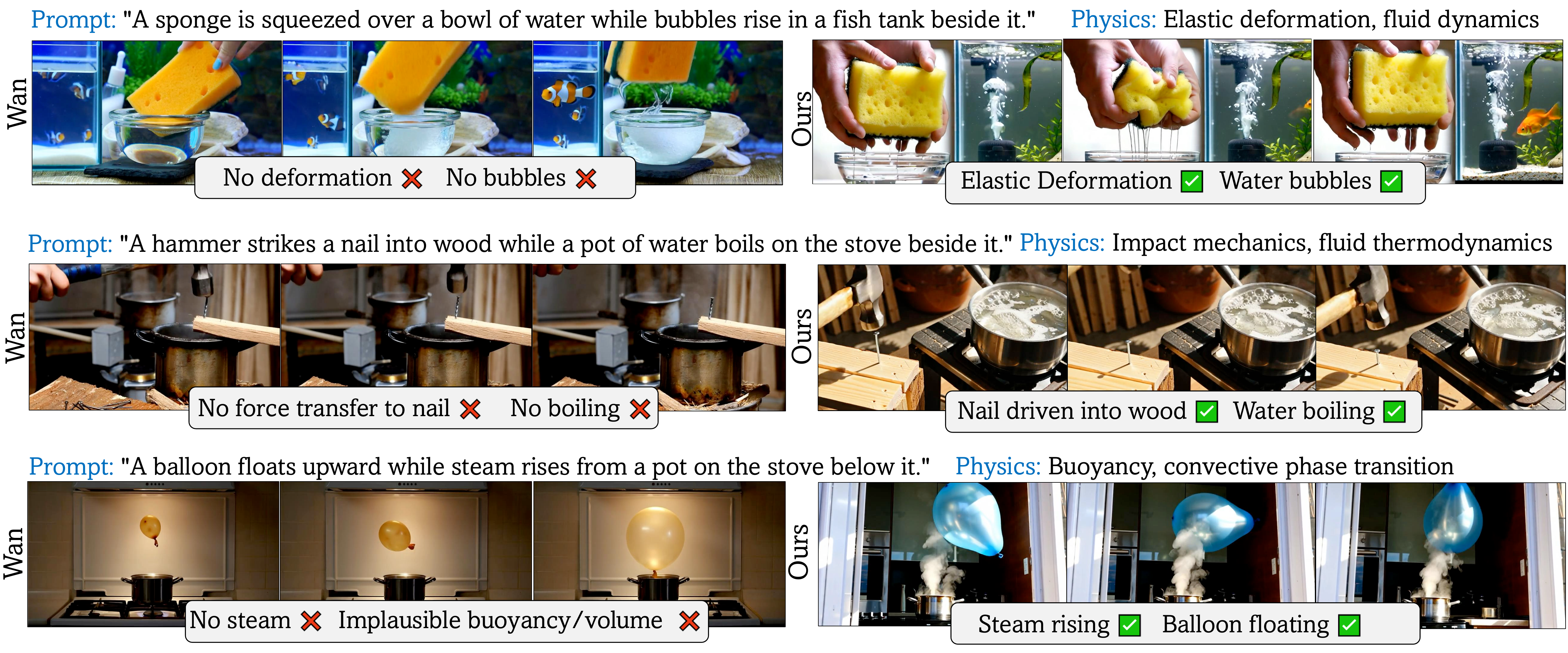}
    \end{adjustbox}
    \vspace{-17pt}
    \caption{\textbf{\texttt{HiPhy} enables generation of physically plausible videos for scenes with multiple concurrent physical principles.} Wan (left) collapses on multi-principle prompts, failing to generate coherent motion or dropping a physical principle entirely, such as missing \textit{boiling} or \textit{steam} motions. Instead of collapsing physical plausibility into a single score, \texttt{HiPhy} (right) aligns generation with a hierarchical reward that tracks the temporal sub-stages of each principle as an independent learning stream, jointly recovering all concurrent processes with correct temporal progression.}
    \label{fig:teaser}
\end{figure}

\maketitle
\begin{abstract}

Video generation models have achieved remarkable visual fidelity and have strong potential to become general-purpose world simulators. Despite this progress, they still fail to generate videos which adhere to laws of physics. The problem becomes even more apparent in realistic settings where multiple physical principles must work together within the same video; for example, \textit{`a balloon floating upward while steam rises from a pot'} requires buoyancy and fluid dynamics to unfold coherently and simultaneously. Yet existing methods largely ignore multi-principle interactions, focusing on a single principle per video. We propose HiPhy (\textbf{Hi}erarchical \textbf{Phy}sical Alignment), a reinforcement learning framework that grounds video generation in physical laws through a dual-level objective: locally enforcing the temporal dynamics of individual physical principles, and globally ensuring the physical and semantic coherence of the entire scene. To support multi-principle generation, we construct a 50K-prompt dataset and introduce a prompt benchmark  \textit{MultiPhyBench},  spanning a diverse range of co-occurring physical events.  Our experiments show that HiPhy significantly outperforms prior methods and baselines, improving physical commonsense and semantic alignment significantly across various benchmarks, with the largest gains on scenes involving multiple concurrent physical principles where competing methods degrade most sharply.

\end{abstract}

\section{Introduction}

\begin{wrapfigure}{r}{0.5\textwidth}
	\centering
	\includegraphics[width=\linewidth]{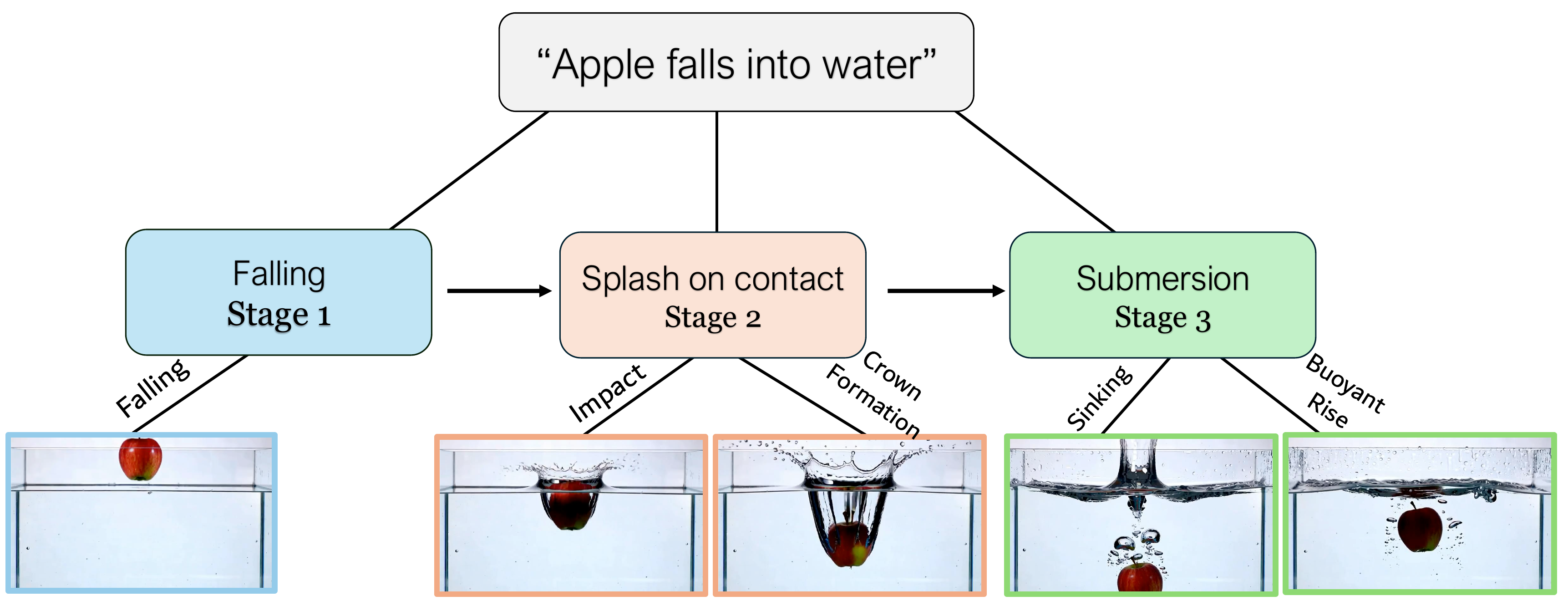}
	\caption{\texttt{HiPhy} explicitly decomposes complex physical events (e.g., an apple falling into water) into temporally ordered sub-stages. This allows us to recursively score the chronological progression, completeness, and alignment of the underlying physics.}
	\label{fig:motivation}
\end{wrapfigure}Video diffusion models \cite{lipman2022flow, song2020denoising} have made remarkable progress in producing high-resolution, temporally coherent videos \cite{wan2025wan, yang2024cogvideox}. This rapid progress has made video generation models promising candidates for general-purpose world simulators that are capable of synthesizing diverse physical scenarios on demand for embodied AI, robotics simulation, scientific visualization, and counterfactual reasoning ~\cite{kang2024far, li2025pisa, kruger2024game, ali2025world}.

Despite their visual fidelity, current video generators routinely produce clips that violate elementary physical laws, such as momentum, fluid dynamics, and material properties \cite{li2025pisa, kang2024far}. We argue this reflects a deeper representational gap: because models are trained merely to maximize frame likelihood given text \cite{ho2022video, ho2022imagen}, they reproduce plausible patterns from their vast training data without reasoning about the underlying mechanisms \cite{motamed2026generative, geirhos2020shortcut}. Consider a prompt \emph{``an apple falls into water,''} which entails two co-occurring physical principles: rigid-body and fluid dynamics (see Fig. \ref{fig:motivation}). We find this representational gap manifests in two consistent failure modes on such compound prompts: (1) \textbf{Principle Omission}, where models ignore secondary physics, for instance, the apple drops realistically but the water remains undisturbed; and (2) \textbf{Temporal Shortcutting}, where unfolding dynamics are collapsed into a single visual snapshot, such as rendering the splash as a static spraying effect instead of progressing through \emph{impact $\rightarrow$ crown formation}. Importantly, these failures do not reflect a deficit in visual capability; given sufficiently detailed text, modern video models generate individual physical principles with striking fidelity. The core challenge is finding a systematic mechanism to elicit this latent knowledge along two dimensions: the presence of each principle and its temporal structure. This motivates \texttt{HiPhy}'s dual-level design: per-principle supervision targets \emph{Principle Omission}, while fine-grained sub-stage reward design addresses \emph{Temporal Shortcutting}. %

Existing efforts to inject physical knowledge into video models largely rely on 3D engines~\cite{liu2024physgen, lv2024gpt4motion, chen2025fluid}, fine-tuning~\cite{cai2025phygdpo, wang2025prophy, gupta2024photorealistic, wang2024worlddreamer}, or inference-time guidance~\cite{yang2025vlipp, huang2025planning, xue2025phyt2v, liu2022compositional, shaulov2025flowmo, jang2026self}. While 3D engines can simulate concurrent dynamics, they are computationally heavy and constrained to rigid domains. Conversely, learning-based and guidance methods offer flexibility but typically supervise only a single physical principle per video. Consequently, they lack mechanisms to ensure compound events are jointly satisfied. Empirically, we observe these latter methods degrade sharply as concurrent processes increase, often dropping one principle entirely.

In this paper, we propose \texttt{HiPhy} (\textbf{Hi}erarchical \textbf{Phy}sical Alignment), a reinforcement learning framework that closes the previously mentioned gap by formulating compound physical events as a hierarchical alignment problem. By optimizing each physical principle locally, \texttt{HiPhy} enables physically grounded video generation in concurrent, multi-principle scenes.  Our key insight is that physical plausibility is fundamentally both a temporal and a compositional property. While modern video generators struggle to reason about the temporal progression of physical events, leading to \emph{Principle Omission} and \emph{Temporal Shortcutting}, we model multi-principle scenes as compositions of independent, temporally structured physical principles. Consequently, we introduce a novel training objective that explicitly enforces stage-level supervision across each concurrent physical principle. \texttt{HiPhy} mirrors this insight through a hierarchical alignment framework. At the foundational level, we explicitly optimize for the chronological progression of each physical event by breaking it into smaller stages. At the same time, we align these stages with global video coherence and semantic alignment objectives, ensuring that multi-principle interactions compose naturally while maintaining semantic and visual coherence. Crucially, when a prompt contains multiple concurrent physical events, \texttt{HiPhy} treats each as an independent learning stream, preventing high-variance signals from washing out the gradient. Our contributions are summarized as follows:
\vspace{-6pt}

\begin{itemize}
    \item   We introduce \texttt{HiPhy}, a reinforcement-learning framework that hierarchically grounds video generation in physical principles. Unlike prior methods that target a single physical principle per video, \texttt{HiPhy} is the first framework that jointly handles multiple concurrent events.
    \vspace{-5pt}
\item  \texttt{HiPhy}   is a plug-and-play post-training framework that integrates with any text-to-video model without architectural modification both open-source (Wan2.1) and black-box (VEO3) backbones.
\vspace{-5pt}

\item \texttt{HiPhy} improves physical commonsense by up to 44\% and semantic alignment by up to 80\% over prior physics-aware baselines.
 \vspace{-4pt}
\item  We curate a 50K-prompt training set and a 1K-prompt benchmark spanning multiple concurrent physical principles, each annotated with per-principle sub-process decompositions. We will share our data, code, and checkpoints publicly.

\end{itemize}

\section{Related Work}
\vspace{-7pt}
\noindent\textbf{Physically Plausible Video Generation}
One line of work fine-tunes video backbones on large physically annotated datasets~\cite{cai2025phygdpo, wang2025prophy, gupta2024photorealistic, wang2024worlddreamer}. Because such models lack explicit physical knowledge, generation quality remains tightly coupled to training-set coverage and rarely transfers to the diversity of real-world scenarios~\cite{motamed2026generative, kang2024far}. A second line injects physical knowledge through 3D engines such as Blender or Unreal, depth or edge predictors, or auxiliary 3D objectives ~\cite{liu2024physgen, lv2024gpt4motion, kupyn2025epipolar}, and graphics rendering pipelines~\cite{an2026vggrpo}. These methods are precise but constrained to fixed physical categories and incur heavy computational cost, particularly for phenomena such as fluids or steam~\cite{chen2025fluid}. Additionally, some works introduce auxiliary signals at inference time, including coarse intermediate representations~\cite{yang2025vlipp, huang2025planning}, off-the-shelf verifiers and judges~\cite{xue2025phyt2v, liu2022compositional}, and test-time optimization~\cite{shaulov2025flowmo, jang2026self}. Verifier-based methods either require large memory budgets or incur long inference times, and test-time optimization typically refines trajectories without addressing implausible initial plans. While prior work holistically supervises a single physical principle per video, \texttt{HiPhy} is the first framework to enforce temporal progression in multi-principle scenes via a hierarchical stage-level objective.

\vspace{-5pt}
\noindent \textbf{RL-based Video Generation}
Prior work adapts preference learning to video along several axes. VideoDPO~\cite{liu2025videodpo} extends DPO to text-to-video diffusion with an omni-preference objective balancing visual quality and text relevance. Flow-DPO \cite{liu2025improving} aligns flow-matching video generators using a multi-dimensional reward, improving motion smoothness and prompt adherence over SFT and alternative RLHF variants. More fine-grained supervision has also been explored via DenseDPO \cite{wu2025densedpo}, which targets temporal locality by densifying video preferences across time. Complementary directions include GRPO variants tailored to video control, such as DanceGRPO \cite{xue2025dancegrpo} for learning temporally coherent policies. Additionally, methods such as \cite{wu2026phyprompt, chen2026hierarchical} provide global supervision over state properties. Unlike these methods, \texttt{HiPhy} is the first to handle multi-principle physical scenes by integrating a hierarchical objective that targets per-principle optimization in its training.
\vspace{-9pt}

\section{Background}
\vspace{-6pt}
\noindent\textbf{Group Relative Policy Optimization}
GRPO~\cite{shao2024deepseekmath} is a reinforcement learning framework that computes advantages through group-based normalization of sampled outputs' rewards, eliminating the need for a separate critic network. Given a query $q$, the method samples $G$ responses $o_1, \ldots, o_G$ from the old policy $\pi_{\theta_{\text{old}}}$ and updates the current policy $\pi_\theta$ to maximize $\mathcal{J}_{\text{GRPO}}(\theta)$ given as:
\begin{equation}
\mathbb{E}_{q, \{o_i\}} \Bigg[ \frac{1}{G} \sum_{i=1}^{G} \min \left( \frac{\pi_\theta(o_i|q)}{\pi_{\theta_{\text{old}}}(o_i|q)} A_i, \right. \\
\left. \text{clip}\left( \frac{\pi_\theta(o_i|q)}{\pi_{\theta_{\text{old}}}(o_i|q)}, 1\!-\!\epsilon, 1\!+\!\epsilon \right) A_i \right) - \beta\, \mathbb{D}_{\text{KL}}(\pi_\theta \,\|\, \pi_{\text{ref}}) \Bigg]
\label{eq:grpo}
\end{equation}

Advantages $A_i$ are computed by normalizing rewards $r_i$ within the group:
\begin{equation}
A_i = \frac{r_i - \operatorname{mean}(r_{1:G})}{\operatorname{std}(r_{1:G})}.
\end{equation}
The standard GRPO objective (Eq.~\ref{eq:grpo}) maximizes the expected reward, with advantage normalization that upweights samples with higher-than-average reward within the group. While effective for single-reward optimization, this formulation collapses distinct reward combinations into identical advantages in multi-reward optimization~\cite{liu2026gdpo}.
\vspace{-5pt}
\section{Method}
\vspace{-2pt}

Given a user prompt $q$ describing $n$ concurrent physical principles and a frozen text-to-video model $\mathcal{G}$, \texttt{HiPhy} learns a policy $\pi_\theta$ such that the video $v = \mathcal{G}(p)$ generated from $p \sim \pi_\theta(\cdot \mid q)$ depicts all $n$ principles with the correct temporal progression of their sub-stages. Two insights shape this design: \textbf{(i)}~Modern video models already possess strong visual priors over individual physical phenomena but lack explicit temporal structure for how a physical event should unfold; surfacing this structure at the prompt level is therefore more sample-efficient than retraining the backbone,  \textbf{(ii)}~Generating complex physical scenes demands compositional reasoning; a video may generate one physical principle faithfully yet fail to compose multiple principles together. We train a reasoning policy $\pi_\theta$ via GRPO~\cite{shao2024deepseekmath} using a hierarchical compositional objective. By explicitly isolating each concurrent physical principle and enforcing the temporal progression of its individual sub-stages, our policy guarantees the structural integrity of compound scenes.

\subsection{Multi-Principle Dataset Curation}
\label{sec:dataset}
We construct a 50K-prompt training dataset and introduce MultiPhyBench, a 1000-prompt evaluation suite in which each prompt describes up to three simultaneous physical events. To generate these, we extract single-principle processes and their domains from VideoPhy2~\cite{bansal2025videophy} and WISA-80k~\cite{wang2025wisa}, sample combinations of 1--3 events, and use an LLM~\cite{yang2025qwen3} to filter out implausible co-occurrences. For each valid combination, we generate a unified video prompt and a chronologically ordered hierarchical tree for each constituent principle (Fig.\ref{fig:motivation}). These hierarchical annotations serve as pseudo-labels for our local reward (Sec.~\ref{sec:reward}). Full curation details are provided in the SM \ref{sec:appendix_dataset}.

\subsection{HiPhy: Hierarchical Physical Alignment}
\label{sec:reward}

\begin{wrapfigure}{r}{0.6\textwidth}
	\centering
	\includegraphics[width=\linewidth]{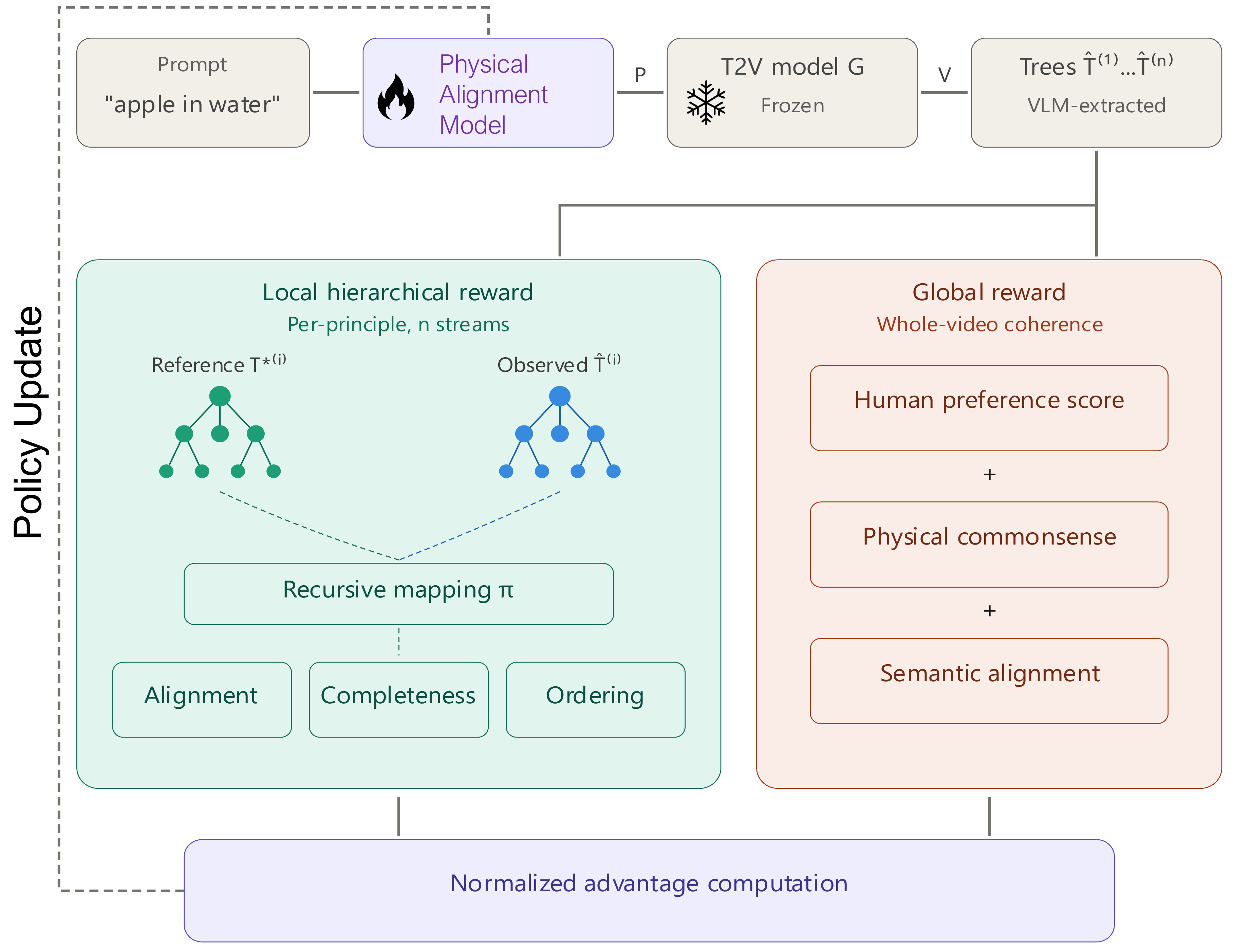}

\caption{Overview of the \texttt{HiPhy} framework. We introduce a hierarchical optimization framework aimed to elicit physically grounded video generation in scenes with multiple physical principles.}
\label{fig:method}
\end{wrapfigure}We introduce  \texttt{HiPhy}, a hierarchical physical alignment framework that scores the temporal progression of each physical principle in a fine-grained fashion (see Fig. \ref{fig:method} for an overview). 
Consider an input prompt $p$ describing $n$ physical principles $\{p^{(i)}\}_{i=1}^n$. During data curation, each physical principle $p^{(i)}$ is paired with an ordered tree $\mathcal{T}^{*(i)}$ of depth $D^{(i)}$ in which each node represents a stage within that physical principle. For \emph{``an apple falls into water and submerges,''} the outer level nodes are \emph{falling $\rightarrow$ splash on contact $\rightarrow$ submerging}. The stage \emph{splash on contact} then further expands into \emph{impact $\rightarrow$ crown formation} (Fig.\ref{fig:motivation}). Nodes in each level are temporally ordered, meaning that the completion of one node depends on the completion of the nodes before it. During training, we generate a video, $\hat{v}$, using a backbone video diffusion model, $\mathcal{G}$ noted as $\hat{v} = \mathcal{G}(p)$. To evaluate how well the generated video, $\hat{v}$, depicts each physical principle $\{p^{(i)}\}_{i=1}^n$, we extract a similar hierarchical structure as $\mathcal{T}^{*(i)}$ from $\hat{v}$ for each physical principle $p^{(i)}$. This procedure is done in the same way as detailed in SM Sec.\ref{sec:appendix_dataset}. Formally, given a VLM extractor, $\mathcal{R}$, we derive an observed tree, $\hat{\mathcal{T}}^{(i)} = \mathcal{R}(\hat{v}, p^{(i)})$, that decomposes the observed stages in the generated video into a hierarchically ordered tree. 

To measure how plausibly each physical principle, $\{p^{(i)}\}_{i=1}^n$, is depicted in the generated video, $\hat{v}$, we align the reference hierarchically ordered tree $\mathcal{T}^{*(i)}$ with the observed tree  $\hat{\mathcal{T}}^{(i)}$. We align the two trees by calculating a one-to-one mapping $\pi$ using the Hungarian assignment~\cite{kuhn1955hungarian}, applied level by level. For each level $L_j$ in $\mathcal{T}^{*(i)}$, we build a similarity matrix $\mathbf{C_{j}}$, where an element $c_{km}$ in $\mathbf{C_{j}}$ measures how similar is node $k$ in $\mathcal{T}^{*(i)}$ to node $m$ in $\hat{\mathcal{T}}^{(i)}$. Each entry in $\mathbf{C_{j}}$ blends a sentence encoder $\phi$ with the VLM judge $\mathcal{J}_{\text{stage}}$ (for computational efficiency, during training, $\mathcal{J}_{\text{stage}}$ is the same as $\mathcal{R}$).
\begin{equation}
\label{eq:sim}
\mathbf{C}_{km} = \alpha \cdot \cos~\!\bigl(\phi(s^*_k), \phi(\hat{s}_m)\bigr) + (1-\alpha) \cdot \mathcal{J}_{\text{stage}}~\!\bigl(s^*_k, \hat{s}_m\bigr),
\end{equation}
where the cosine term handles paraphrase and the VLM judge catches cases where wording is similar but the underlying event differs. If node $a$ in $\mathcal{T}^{*(i)}$ has sub-stages in level $L_{j+1}$, we recursively build $\mathbf{C_{j+1}}$ for the nodes in $L_{j+1}$ and calculate our mapping $\pi$ for each level $L_{j}$ from its associated $\mathbf{C_{j}}$. Please refer to SM \ref{alg:align} for full pseudocode. 

Using the mapping $\pi$ and the similarity matrices, we evaluate how well the video depicts the physical principle along three dimensions: \emph{Alignment} $r_{\text{align}}^{(i)}$, \emph{completeness} $r_{\text{comp}}^{(i)}$, and \emph{ordering} score $r_{\text{order}}^{(i)}$.

First, let $t^{(i)}$ denote the total number of leaf nodes (the most granular sub-stages) in the reference tree, and let $t^{(i)}_{\text{matched}}$ be the subset of successfully matched leaves. A leaf is in $t^{(i)}_{\text{matched}}$ if its similarity score and all of its parent nodes in the similarity matrix $\mathbf{C}$ are above a threshold $\tau$. This logical dependency ensures that if a high-level stage fails to generate (e.g., missing the \emph{splash}), all of its dependent sub-stages (e.g., \emph{crown formation}) are strictly treated as missing. \textit{Alignment} calculates the average similarity of the successfully matched leaves to assess how well the observed stages in $\hat{v}$ match the expected reference stages in $\mathcal{T}^{*(i)}$, while \textit{completeness} calculates the fraction of expected leaves that passed the threshold to assess how much of the physical process was captured:
\begin{equation}
r_{\text{align}}^{(i)} = \frac{1}{t^{(i)}}\!\!\sum_{k \in t^{(i)}_{\text{matched}}}\!\! \mathbf{C}_{k,\pi(k)}, \quad
r_{\text{comp}}^{(i)} = \frac{1}{t^{(i)}}\!\!\sum_{k \in t^{(i)}_{\text{matched}}}\!\! \mathbf{1}\!\left[\mathbf{C}_{k,\pi(k)} \geq \tau\right].
\end{equation}
To ensure the generated sub-stages unfold in the correct chronological sequence, we compute an \emph{ordering} score $r_{\text{order}}^{(i)}$. 
We evaluate this locally for each successfully matched parent node $\mathbf{u} \in \mathcal{N}^{(i)}$. Letting $\mathcal{M}_\mathbf{u} = (m_1, \ldots, m_{|\mathcal{M}_\mathbf{u}|})$ be its sequence of matched children ordered by their expected reference position, the local ordering score $O_\mathbf{u}^{(i)}$ calculates the fraction of these children that appear in the correct relative temporal sequence ($\succ_\mathbf{u}$) within the generated video:
$$O_\mathbf{u}^{(i)} = \frac{1}{|\mathcal{M}_\mathbf{u}|-1}\sum_{k=2}^{|\mathcal{M}_\mathbf{u}|} \mathbf{1}\!\left[\pi(m_k) \succ_{\mathbf{u}} \pi(m_{k-1})\right],$$
where $\pi(m_k) \succ_\mathbf{u} \pi(m_{k-1})$ indicates that the observed sibling matched to $m_k$ appears later than the one matched to $m_{k-1}$, and $O_\mathbf{u}^{(i)} = 1$ when $|\mathcal{M}_\mathbf{u}| \leq 1$. The principle-level ordering score is then the geometric mean across all matched internal nodes, given by $r_{\text{order}}^{(i)} = \bigl(\prod_{\mathbf{u} \in \mathcal{N}^{(i)}} O_\mathbf{u}^{(i)}\bigr)^{1/|\mathcal{N}^{(i)}|}$. Finally, the hierarchical reward for the physical principle is the mean of these three sub-scores:

\begin{equation}
\label{eq:local_hierc}
R_{\text{hier}}^{(i)}(\hat{v}) = \frac{1}{3}\bigl(r_{\text{comp}}^{(i)} + r_{\text{align}}^{(i)} + r_{\text{order}}^{(i)}\bigr).
\end{equation}
Notably, if a physical principle lacks hierarchical depth ($D^{(i)} = 1$), the recursion bottoms out after a single assignment at the root, naturally recovering a standard flat-stage reward formulation as a special case.

\noindent\textbf{Global Video Reward.} The local reward captures fine-grained temporal progression within each process but does not guarantee that the video is semantically faithful to the prompt, visually coherent, or consistent with broader physical commonsense. We introduce a global reward term $R_{\text{global}}$ with three complementary terms. First, HPSv2~\cite{wu2023human} captures visual quality and aesthetic alignment, averaged over $T$ uniformly sampled frames to obtain $R_{\text{HPS}}$. Second, a physics VLM judge $\mathcal{J}$ (VideoPhy2~\cite{bansal2025videophy}) rates physical commonsense $R_{\text{PC}}$, assessing whether the primary action follows real-world physical rules, and semantic alignment $R_{\text{SA}}$, measuring whether events described in $q$ actually appear in $\hat{v}$. All three terms are linearly normalized to $[0, 1]$ and combined into a single scalar:
\begin{equation}
R_{\text{global}}(\hat{v}, q) = w_h \tilde{R}_{\text{HPS}} + w_p \tilde{R}_{\text{PC}} + w_s \tilde{R}_{\text{SA}},
\end{equation}
Because all three terms characterize whole-video quality, we treat $R_{\text{global}}$ as a single reward stream. Together, $R_{\text{PC}}$ guards against gross physical implausibility that local rewards may overlook, $R_{\text{SA}}$ anchors generation to prompt semantics, and $R_{\text{HPS}}$ favors visually coherent outputs.

\noindent\textbf{Training}
For a prompt with $n$ principles, we sample $G$ rollouts $\{\hat{v}^{(j)}\}_{j=1}^{G}$, yielding $n$ hierarchical rewards $R_{\text{hier}}^{(i,j)}$ and one global reward $R_{\text{global}}^{(j)}$. To prevent high-variance streams from dominating the GRPO~\cite{shao2024deepseekmath} gradient, we independently normalize each stream within the group~\cite{liu2026gdpo}:
\begin{align}
A_{\text{hier}}^{(i,j)} = \frac{R_{\text{hier}}^{(i,j)} - \mu_i}{\sigma_i + \epsilon}, \quad
A_{\text{global}}^{(j)} = \frac{R_{\text{global}}^{(j)} - \mu_g}{\sigma_g + \epsilon},
\end{align}
where $\mu$ and $\sigma$ denote the per-stream mean and standard deviation across the $G$ rollouts. Summing these unit-variance local streams directly would scale the overall variance by $\mathcal{O}(n)$, disproportionately skewing the gradient for prompts with more principles. To maintain consistent variance regardless of $n$, we rescale the local sum by $1/\sqrt{n}$:
\begin{equation}
\label{eq:norm}
A_{\text{sum}}^{(j)} = A_{\text{global}}^{(j)} + \frac{1}{\sqrt{n}}\sum_{i=1}^{n} A_{\text{hier}}^{(i,j)}.
\end{equation}
Finally, we apply batch-wise normalization $\hat{A}^{(j)} = (A_{\text{sum}}^{(j)} - \mu_b)/(\sigma_b + \epsilon)$ to absorb residual scale variation from across-prompt heterogeneity~\cite{liu2026gdpo}. This $\hat{A}^{(j)}$ serves as the final advantage in the clipped GRPO objective, utilizing a per-token KL penalty against a frozen reference policy.
\begin{figure*}[t!]
\centering
\includegraphics[width=0.95\textwidth]{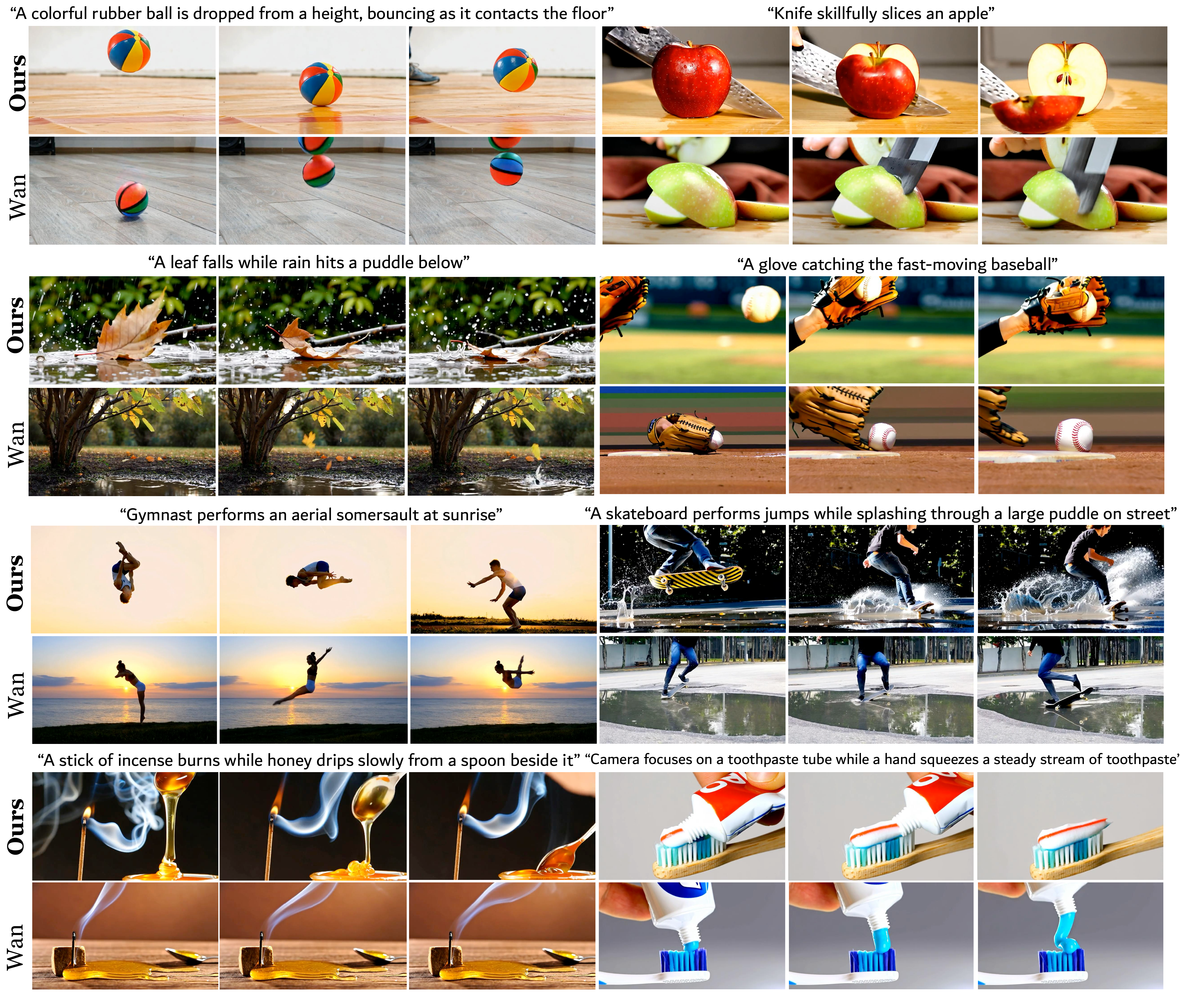}
    \caption{\textbf{Qualitative results.} \texttt{HiPhy} (top row) produces coherent motion and correct temporal progression on prompts spanning single- and multi-principle physical scenes, while Wan2.1 (bottom row) frequently drops a principle, breaks contact dynamics, or generates static scenes. }
\label{fig:qual_res}
\end{figure*}

\section{Experiments}
\label{sec:experiments}
We showcase the effectiveness of our method on state-of-the-art T2V models Wan2.1~\cite{wan2025wan}, as well as black-box T2V model VEO3~\cite{DeepMindVeo3TechReport2025}. We compare our method against finetuning methods \cite{wang2025wisa, zhang2025videorepa}, inference optimization \cite{jang2026self}, prompt refinement methods \cite{xue2025phyt2v, singh2025openai}. For video generation, we utilize Wan2.1 \cite{wan2025wan} as our backbone and Qwen2.5-7b-Instruct~\cite{yang2025qwen3} as our alignment model. In Eq. \ref{eq:sim}, $\alpha$ is set $0.5$ and $\tau$ is set $0.65$ through a hyperparameter search. The sentence encoder $\phi$ is SBERT ~\cite{reimers2019sentence} and the VLM judge is Qwen2.5-VL-7B~\cite{bai2023qwen}.

\subsection{Experimental Setup}
 All experiments are conducted on two NVIDIA H200 GPUs. We first perform supervised finetuning for 50 iterations, followed by GRPO training for approximately 2000 iterations. The learning rate starts from $2\times10^{-5}$ for SFT and $2\times10^{-7}$ during GRPO optimization, with $w_h$, $w_p$, and $w_s$ set to $0.33$. Our dataset comprises of $\approx50K$ prompts with 1--3 concepts equally distributed. Validation and test data are \%10 and \%5 of the data, respectively.  We use a group size of 4 and a training batch size of 8, optimizing with Adam optimizer \cite{kingma2014adam}, and TRL \cite{vonwerra2022trl} library. Unless otherwise specified, we adopt a guidance scale of 6, with 40 inference steps and 81 output frames per video.
\subsection{Qualitative Comparisons}
Figure~\ref{fig:qual_res} illustrates the diverse physical principles our method handles compared to the Wan~\cite{wan2025wan} backbone. Scenes span rigid-body dynamics (\textit{bouncing ball, sliced apple, catching a baseball}), fluid-rigid interactions (\textit{leaf falling into a puddle}), athletic motion (\textit{somersault, skateboarding}), and concurrent multi-principle events (\textit{smoke} beside \textit{dripping honey}, \textit{squeezing toothpaste}). In these scenarios, Wan frequently fails: balls bounce implausibly or multiply randomly, leaves disappear into water rather than floating, and athletic motions collapse. Conversely, our method generates each process with accurate temporal progression.

\begin{figure*}[t!]
\centering
\includegraphics[width=\textwidth]{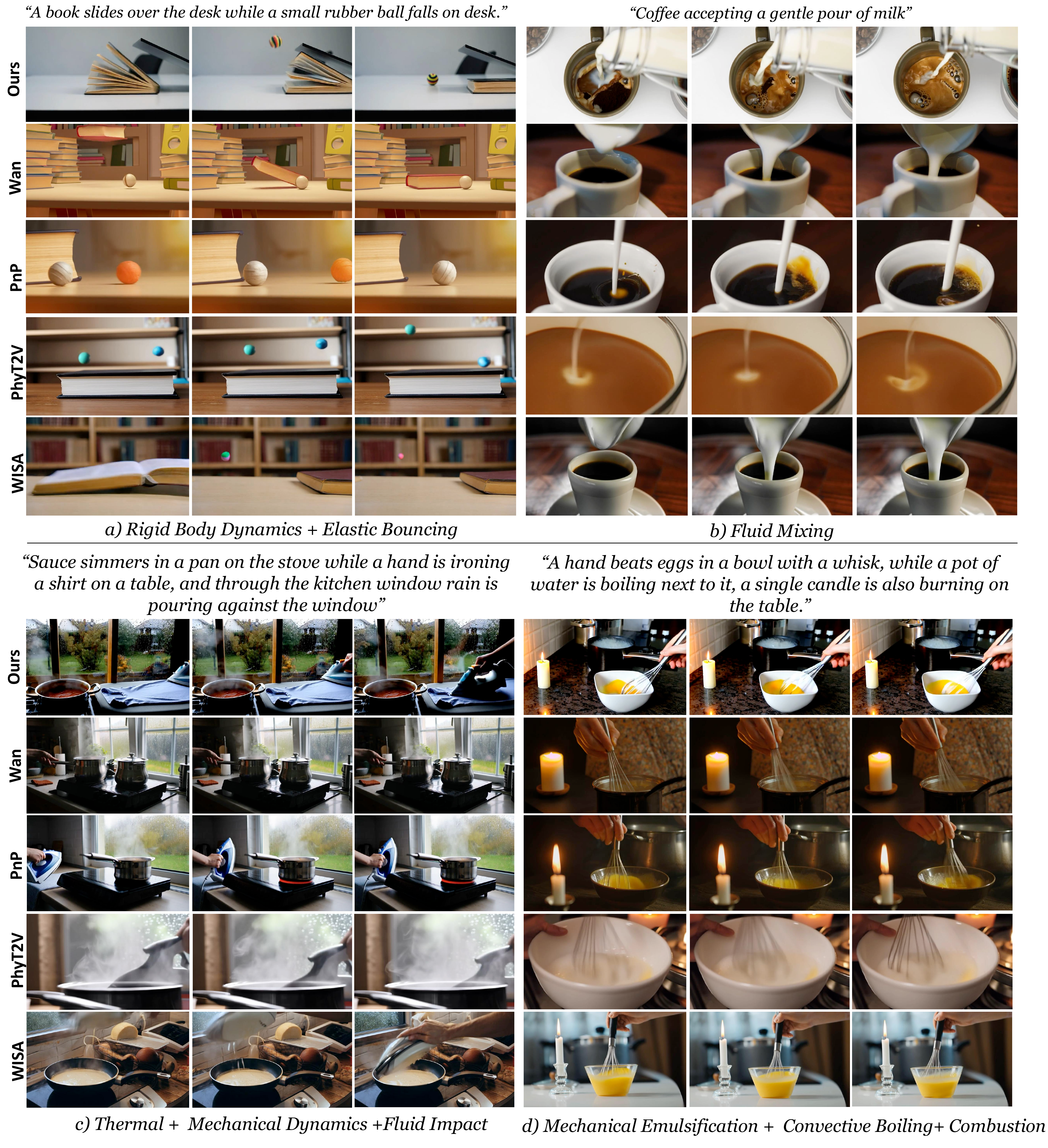}
    \vspace{-1em}
    \caption{\textbf{Qualitative comparisons.} As the number of physical principles per prompt grows, baselines exhibit consistent failure modes, incoherent motion, hallucinated objects, or skipping a principle entirely, and these compound on three-principle prompts (bottom) where multiple processes are dropped simultaneously. \texttt{HiPhy} (top row of each block) renders each constituent process coherently without degrading visual quality.}
    \vspace{-25pt}
\label{fig:qual}
\end{figure*}

Figure~\ref{fig:qual} compares our method against Wan and three physics-aware baselines: inference-time optimization (Predict-and-Perturb~\cite{jang2026self}), iterative LLM refinement (PhyT2V~\cite{xue2025phyt2v}), and finetuning (WISA~\cite{wang2025wisa}). The results reveal consistent baseline failure modes that compound as the physical event count increases. On two-principle prompts like ``\textit{book sliding while a ball falls}'', Fig.\ref{fig:qual}, \textbf{a}, baselines struggle with coherent interaction: Wan and PnP omit the \textit{sliding book}, PhyT2V hallucinates extra objects, and WISA produces static scenes. Similarly, in Fig.\ref{fig:qual}, \textbf{b}, gradual integration is often skipped. This performance gap widens on three-principle prompts (bottom). Baselines routinely drop key elements, such as the \textit{simmering sauce} or \textit{rain} in Fig.\ref{fig:qual}, \textbf{c}, or the \textit{boiling water} and \textit{whisking action} in Fig.\ref{fig:qual}, \textbf{d}. In contrast, our method coherently generates all constituent processes without degrading overall video quality, which we attribute to optimizing each process as a separate local reward stream during training.

\subsection{Quantitative Experiments}
\begin{table*}[t]
\centering
\caption{Comprehensive evaluation  across VideoPhy2, PhyWorldBench, Multi-concept, and VBench, and ablation experiments over different reward terms}
\label{tab:main_eval}
\resizebox{\textwidth}{!}{
\setlength{\tabcolsep}{3pt} 
\begin{tabular}{l|ccc|ccc|ccc|cc}
\toprule
 & \multicolumn{3}{c|}{\textbf{VideoPhy2}} & \multicolumn{3}{c|}{\textbf{PhyWorldBench}} & \multicolumn{3}{c|}{\textbf{MultiPhyBench}} & \multicolumn{2}{c}{\textbf{VBench}} \\
 \cmidrule(lr){2-4} \cmidrule(lr){5-7} \cmidrule(lr){8-10} \cmidrule(l){11-12}
Method & PC$\uparrow$ & SA$\uparrow$ & PC \& SA $\uparrow$ & PC$\uparrow$ & SA$\uparrow$ & PC \& SA $\uparrow$ & PC$\uparrow$ & SA$\uparrow$ & PC \& SA$\uparrow$ & Dyn. $\uparrow$ & App.$\uparrow$ \\
\midrule
Wan2.1    & 49.10 & 43.60 & 30.5   & 47.7 & 40.9 & 23.7          & 53.0 & 35.0 & 27.0 & 0.808 & \underline{0.573}\\
PnP       & \underline{80.1} & \underline{68.0} & \underline{60.14} & 56.1 & 48.5 & 31.2 & \underline{57.0} & \underline{37.0} & \underline{29.0} & 0.844 & 0.549\\
VideoRepa & 70.9 & 53.5 & 43.8     & 51.7 & 33.3 & 19.3 & 49.0 & 22.0 & 15.0 & 0.863 & 0.550\\
PhyT2V    & 79.0 & 65.0 & 55.0& 40.3 & 47.5 & 22.7& 50.0 & 26.0 & 15.0 & 0.872 & 0.500\\
WISA      & 68.0 & 59.7 & 44.3 & 56.3 & 50.7 & 30.7 & 52.5 & 43.0 & 24.5 & 0.859 & 0.567\\

GPT-5 &  69.2 & 60.9 & 43.1 & \underline{56.9} & \underline{52.5} & \underline{31.8} & 51.5 & 44.0 & 24.5 & \underline{0.882} & 0.531\\

\rowcolor[HTML]{F0F5FF} HiPhy (Ours) &\textbf{ 88.6} & \textbf{79.2} & \textbf{68.8} & \textbf{74.9} & \textbf{76.3} & \textbf{61.6} & \textbf{76.7} & \textbf{68.85} & \textbf{52.7} & \textbf{0.905} & \textbf{0.592}\\
\midrule
Qwen 2.5 & 60.4 & 62.9 & 41.5 &58.6 & 58.3 & 44.5& 50.3 & 39.2 & 19.1 & 0.874 & 0.517 \\
 HiPhy (global)    & \underline{72.7} & \textbf{75.3} & \underline{56.3} & \underline{62.3} & \textbf{64.3} & \underline{54.3} & 61.0 & 53.0 & 40.0 & 0.871 & \textbf{0.576}\\
HiPhy (local, sum) & 72.2 & 70.1 & 48.5& 61.9 & 57.3 & 37.1 & \underline{67.0} & \underline{57.9} & \underline{42.4} & 0.865 & 0.556 \\
HiPhy (local, normalized) & \textbf{79.2} & \underline{70.4} & \textbf{57.3} & \textbf{65.6} & \underline{60.8} & \textbf{52.5}& \textbf{68.5} & \textbf{58.3} & \textbf{44.5} & \textbf{0.893} & 0.567 \\

\bottomrule
\end{tabular}
}

\end{table*}
We evaluate our method across multiple physical benchmarks such as Videophy2 and PhyWorldBench~\cite{bansal2025videophy, gu2025phyworldbench}. We assess physical commonsense (PC) and semantic alignment (SA) using Gemini 3 Flash \cite{team2024gemini}, which supports higher frame rates. Additionally, we report a ``PC \& SA'' metric that measures the fraction of videos that are simultaneously physically grounded and semantically aligned. To assess robustness beyond physics-specific benchmarks, we also compare against baselines on VBench~\cite{huang2024vbench}, summarizing its orthogonal dimensions into two axes: \textit{Dynamics} (subject and background consistency, motion smoothness, dynamic degree) and \textit{Appearance} (imaging and aesthetic quality). Table~\ref{tab:main_eval} reports results across two single-principle benchmarks (VideoPhy2~\cite{bansal2025videophy} and PhyWorldBench~\cite{gu2025phyworldbench}) and our curated multi-principle benchmark of 1K prompts containing concurrent physical principles. \texttt{HiPhy} achieves the strongest performance on all physics benchmarks. Our method significantly outperforms the best baselines on VideoPhy2 (+10\% physical, +16\% alignment) and PhyWorldBench ($\approx$ +30\% physical, +45\% semantic), demonstrating the strength of our hierarchical reward. While generating concurrent physical events remains challenging across all methods, our per-principle hierarchical optimization preserves distinct learning signals, maintaining the largest margin on the multi-principle benchmark.
Finally, VBench \cite{huang2024vbench} evaluations confirm that physics-aware training preserves general video quality. Both \textit{Dynamics} and \textit{Appearance} match or exceed the Wan2.1 baseline, as our global reward terms effectively anchor overall coherence.

\subsection{Human Evaluation.}
We conducted a user study ($n=50$) with 30 prompts to evaluate \textit{physical commonsense} and \textit{semantic alignment} using a 1--5 Likert scale. 
\begin{wraptable}{r}{0.31\textwidth}
\centering
\small
\setlength{\tabcolsep}{4.4pt}
\begin{tabular}{lccc}
\toprule
Method & PC & SA & Principle \\
\midrule
PhyT2V & 2.28 & 2.11 & 2.19 \\
WAN    & 2.76 & 2.19 & 2.40 \\
WISA   & 2.49 & 2.29 & 2.11 \\
PnP    & 3.07 & 2.90 & 2.67 \\
\textbf{Ours}   & \textbf{3.89} & \textbf{4.13} & \textbf{3.81} \\

\bottomrule

\end{tabular}
\caption{User Study.}
\label{tab:user_study}
\end{wraptable}As shown in Table~\ref{tab:user_study}, our method outperforms all baselines across all metrics, with the most significant margin in text alignment, indicating that raters perceive a clear improvement in compositional physical fidelity. A sample survey question is provided in the SM. Furthermore, we analyzed performance on videos containing multiple concurrent physical principles (\textbf{Fig.~\ref{fig:per-process}, D}).

While baselines exhibit weak performance, often failing to depict specific principles entirely within a scene, \texttt{HiPhy} maintains steady, high ratings across all individual events. This consistency validates our per-process reward design, which ensures each physical principle receives a distinct learning signal.

\subsection{Computational Efficiency.} We compare the inference time of our method against prior physics-grounded video generation baselines on a single H200 GPU. PhyT2V \cite{xue2025phyt2v} relies on iterative VLM-based prompt refinement and re-generation, incurring substantial overhead on top of the base video model. PnP \cite{jang2026self} incurs extra computation for its extra optimization steps during inference. WISA \cite{wang2025wisa} introduces moderate cost through fine-tuned LoRA. In contrast, \texttt{HiPhy} runs once before generation and adds only $\sim$2.4 seconds of overhead, making the total inference time effectively identical to the base model while achieving stronger physical grounding.

\vspace{-5pt}
\section{Ablation}
\vspace{-5pt}

\subsection{Reward Composition}
We ablate the contribution of each reward component in Table~\ref{tab:main_eval}. \emph{Qwen 2.5} denotes the base model with no RL training; \texttt{HiPhy} (global) trains with only the global reward; \texttt{HiPhy}  (local, sum) \begin{wraptable}{r}{0.35\textwidth}
\vspace{-1pt}
\centering
\small
\setlength{\tabcolsep}{6pt}
\begin{tabular}{lcc}
\toprule
Method & Time (min) & Overhead \\
\midrule
PhyT2V & 8.79 & $+170$s \\
WISA   & 6.66 & $+36$s  \\
PnP    & 8.86 & $+175$s \\
Wan    & 5.96 & ---     \\
\midrule
\textbf{Ours} & \textbf{6.00} & $\mathbf{+2.4}$\textbf{s} \\
\bottomrule
\end{tabular}
\caption{Computational efficiency comparison. \texttt{HiPhy} achieves minimal overhead.}
\label{tab:time_complexity}
\vspace{-1pt}
\end{wraptable}trains with only the local hierarchical rewards summed as a single scalar; and \texttt{HiPhy}  (local, normalized)  trains the hierarchical reward with normalized streams as in Eq. \ref{eq:norm}. The two reward families turn out to be complementary. Training with only the global reward yields strong semantic alignment and visual quality but weaker physical commonsense, since the global reward provides no fine-grained signal about how individual processes unfold over time. Training with only the local reward improves physical commonsense on multi-process scenes but degrades semantic alignment and dynamic degree, as the policy optimizes for stage-level correctness without anchoring to prompt semantics or visual coherence. Combining both with per-stream normalization recovers the best of both, as shown in \texttt{HiPhy} row, further confirming our design choice.

\subsection{Generalization across video backbones.}
\begin{figure*}[!htb]
\centering
\includegraphics[width=\textwidth]{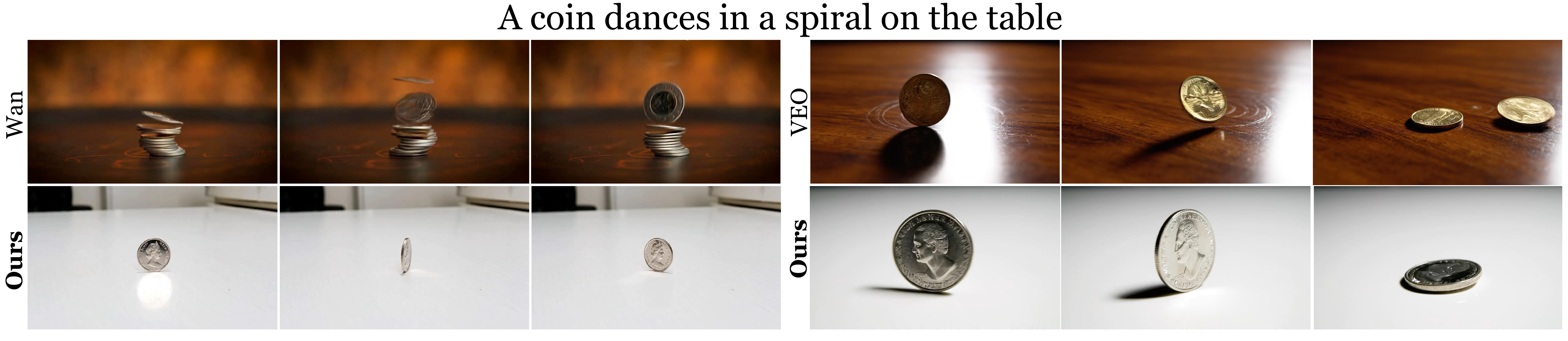}
    \caption{\textbf{Ablation on different backbones.}}
    \vspace{-7pt}
\label{fig:ablation_backbone}
\end{figure*}

Figure~\ref{fig:ablation_backbone} shows that both Wan~\cite{wan2025wan} and VEO~\cite{DeepMindVeo3TechReport2025} produce near-static coins on the prompt \textit{``a coin dances in a spiral on the table,''} despite the motion being well within their generative capacity. Pairing each backbone with \texttt{HiPhy} recovers the spiral trajectory in both cases, supporting our claim that the failure to render physically plausible scenes reflects a representational gap rather than a deficit of visual capability. 
\texttt{HiPhy} transfers across any backbone with sufficient underlying physical priors without retraining.

\subsection{Performance vs. principle count.}

\begin{figure*}[!htb]
\centering
\includegraphics[width=0.9\textwidth]{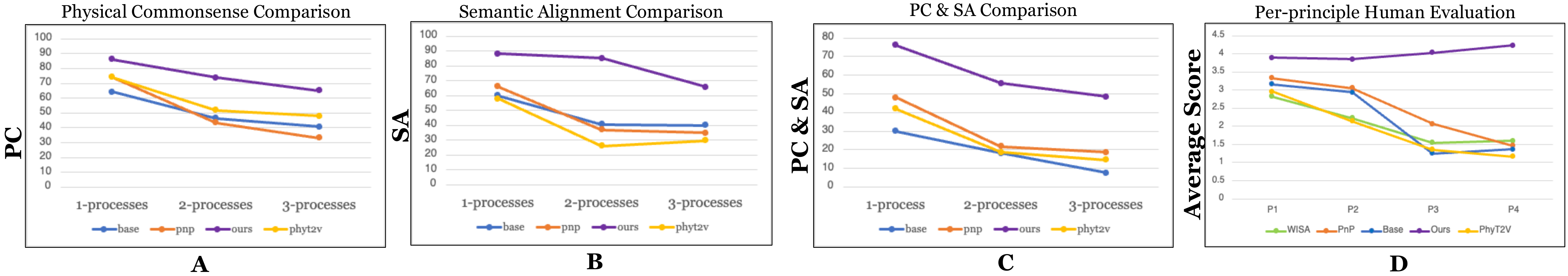}
    \caption{\textbf{Performance scaling and human evaluation.} (A–C) Automatic metrics for PC, SA, and joint PC\&SA across process counts. \texttt{HiPhy} (purple) maintains a significantly wider margin as complexity increases. (D) Averaged human scores per principle (P1–P4). While baselines show constant degradation in multi-event scenes, \texttt{HiPhy} achieves consistent physical fidelity across all concurrent principles.}
    \vspace{-5pt}
\label{fig:per-process}
\end{figure*}

Figure~\ref{fig:per-process} \textbf{(A--C)} illustrates performance as the number of concurrent principles per prompt increases. While all methods degrade with rising complexity, baselines exhibit a sharp collapse; PnP and PhyT2V lose approximately half their PC scores between 1- and 3-process prompts, with joint PC \& SA scores falling below 20. In contrast, \texttt{HiPhy} maintains a joint score of approximately 50, widening the performance gap as complexity grows. We attribute this robustness to our per-process reward stream, which provides distinct learning signals for each event. This prevents the gradient dilution seen in baselines that attempt to collapse multi-process correctness into a single scalar.

\vspace{-8pt}
\section{Limitations and Broader Impacts} \texttt{HiPhy} uses a frozen backbone T2V model, so its fidelity is upper-bounded by the visual capacity of the underlying generator, similar to other existing work \cite{xue2025phyt2v, yang2025vlipp}.  More broadly, physics-aware generation supports applications in education, scientific visualization, and embodied AI, but more plausible video raises the stakes for misinformation and non-consensual content; we build on public backbones and datasets and encourage provenance tools and responsible-release practices.
\vspace{-6pt}

\section{Conclusion} We presented a reinforcement learning approach to physics-grounded video generation that decomposes physical correctness into per-principle local rewards over temporal sub-stages and a global reward for semantic and visual coherence. Across VideoPhy2, PhyWorldBench, and our multi-principle benchmark, \texttt{HiPhy} achieves the strongest physical commonsense and semantic alignment, with the largest margins on prompts containing multiple concurrent principles, where prior methods degrade most sharply. These results suggest that explicit temporal decomposition, combined with a learned refinement signal, is an effective approach to closing the gap between video generation and physical realism.

\section{Acknowledgments}

This work is supported by a Qualcomm Gift and National Science Foundation under Grant No. 2543524. We also gratefully acknowledge fal.ai for providing compute support.

\clearpage

\clearpage
\bibliographystyle{splncs04}
 \bibliography{main}
\clearpage
\appendix

\section*{Table of Contents}
\addcontentsline{toc}{section}{Supplementary Material Table of Contents}
\startcontents[appendix]
\printcontents[appendix]{l}{1}{\setcounter{tocdepth}{2}}

\newpage 

\section{Multi-Process Dataset Curation}
\label{sec:appendix_dataset}
Existing physical video datasets and benchmarks evaluate physical plausibility along a single process per prompt (e.g., a ball rolling down an incline or water pouring into a glass). However, real-world scenes routinely involve multiple physical processes unfolding simultaneously. To model physical realism in multi-process settings, we construct a dataset in which each prompt describes multiple concurrent physical events. The number of processes per prompt, $n$, scales with the video length; for instance, at 81 frames we support up to three simultaneous processes.

\noindent \paragraph{Physical Process Extraction.} We collect single-process physical concepts from existing datasets, including VideoPhy2~\cite{bansal2025videophy} and WISA-80k~\cite{wang2025wisa}. For each source prompt, we use a large language model to extract the underlying physical process along with its physics domain. For instance, from the prompt \textit{``a bowling ball rolls off a table and hits the ground,''} we extract the process \textit{``ball rolling and falling''} with the domain \textit{rigid-body
dynamics}; from \textit{``steam rises from a boiling pot,''} we extract \textit{``water boiling''} with the domain \textit{thermodynamics}.

\paragraph{Combinatorial Sampling and Plausibility Filtering.} We sample subsets of 1, 2, or 3 physical concepts per prompt. To capture the true complexity of real-world scenes, we allow combinations of both inter-domain events (e.g., fluid dynamics paired with rigid-body physics) and simultaneous intra-domain events (e.g., multiple concurrent rigid-body interactions). Because not all sampled combinations describe events that can realistically co-occur within a single scene, we employ an LLM to judge whether a given set of processes is physically compatible, discarding combinations marked as implausible.

\paragraph{Prompt Generation and Stage Annotation.} For each plausible combination, we generate a natural-language video description that coherently integrates all constituent processes within a single scene. In addition, for each physical process, we generate a sequence of chronologically ordered stages that describe the progression of that process over time (e.g., \textit{contact $\rightarrow$ momentum transfer $\rightarrow$ deceleration} for a collision). These stage annotations serve as pseudo-labels references for the local stage-aware reward described in the next section. Figure \ref{fig:motivation} the structure of the resulting annotations.

\section{Hierarchical alignment algorithm}
\label{app:align}

We provide pseudocode for the recursive alignment procedure used to compute the hierarchical reward (Sec.~\ref{sec:reward}). Given a reference tree $\mathcal{T}^{*(i)}$ and an observed tree $\hat{\mathcal{T}}^{(i)}$ extracted by the VLM, the procedure is invoked at the roots and descends both trees in parallel, solving a local assignment problem at each matched node and recursing into matched pairs whose similarity exceeds the threshold $\tau$.
\begin{algorithm}[H]
\caption{\textsc{HiPhyAlign}($s^*, \hat{s}, \Pi, \mathcal{S}, \mathcal{N}$) — recursive tree alignment}
\label{alg:align}
\small
\begin{algorithmic}[1]
\Require reference node $s^*$, observed node $\hat{s}$; accumulators $\Pi$ (assignment), $\mathcal{S}$ (matched-leaf similarities), $\mathcal{N}$ (matched internal nodes)
\If{$s^*$ is a leaf}
  \State $\mathcal{S} \gets \mathcal{S} \cup \{(s^*, \mathbf{C}_{s^*, \hat{s}})\}$
  \State \Return
\EndIf
\State $\mathcal{N} \gets \mathcal{N} \cup \{s^*\}$
\State $S^* \gets \mathrm{ch}(s^*) = (s^*_1, \ldots, s^*_K)$, $\hat{S} \gets \mathrm{ch}(\hat{s}) = (\hat{s}_1, \ldots, \hat{s}_M)$
\State Build $\mathbf{C} \in \mathbb{R}^{K \times M}$ with entries from Eq.~\ref{eq:sim}; pad with zeros to a square matrix
\State $\pi_{\text{local}} \gets \arg\max_\pi \sum_k \mathbf{C}_{k, \pi(k)}$ \Comment{Hungarian assignment}
\For{$k = 1, \ldots, K$}
  \State $\Pi \gets \Pi \cup \{(s^*_k, \hat{s}_{\pi_{\text{local}}(k)})\}$
  \If{$\mathbf{C}_{k, \pi_{\text{local}}(k)} \geq \tau$}
    \State \textsc{HiPhyAlign}$(s^*_k, \hat{s}_{\pi_{\text{local}}(k)}, \Pi, \mathcal{S}, \mathcal{N})$ \Comment{recurse on reliable matches only}
  \EndIf
\EndFor
\end{algorithmic}
\end{algorithm}

The procedure is invoked once per principle as \textsc{HiPhyAlign}$(\mathrm{root}(\mathcal{T}^{*(i)}), \mathrm{root}(\hat{\mathcal{T}}^{(i)}), \emptyset, \emptyset, \emptyset)$, after which $\Pi$ contains the global assignment $\pi$ pairing every reference node with its observed counterpart, $\mathcal{S}$ holds the leaf-level similarities $\mathbf{C}_{k, \pi(k)}$ used by $r_{\text{align}}^{(i)}$ and $r_{\text{comp}}^{(i)}$, and $\mathcal{N} = \mathcal{N}^{(i)}$ is the set of matched internal nodes used by $r_{\text{order}}^{(i)}$. Descendants of a sub-threshold match are never recursed into and count as unmatched: their leaves enter the denominators of $r_{\text{align}}^{(i)}$ and $r_{\text{comp}}^{(i)}$ but not the numerators. The Hungarian step has cost $O((K+M)^3)$ per node; the branching factor at every level is small in our datasets (typically $\leq 5$), so alignment cost is negligible compared to video generation.
\section{More Qualitative }

Please refer to our supplementary webpage (.html) file in our supplementary material for more qualitative examples and video samples mentioned in the main paper.

\section{Extracted Stage validation} To validate that the sub-stages extracted by our pipeline correspond to events that humans actually perceive in the underlying videos, we conducted a small-scale human study on 50 randomly sampled videos spanning six physics domains. Each video was rated by 5 annotators on a 1–5 Likert scale along two axes: how clearly each sub-stage was depicted and whether the sub-stages occurred in the listed chronological order. Table~\ref{tab:human_eval_stage} reports the results. The mean stage-quality rating of 4.21/5 and ordering rating of 4.42/5 indicate that our extracted decompositions are faithful to the videos at both the level of individual sub-stages and their temporal arrangement. Scores are consistent across domains, with optics events scoring lowest (3.88), reflecting the difficulty of extracting discrete temporal stages from scenes dominated by events such as reflection and scattering. Furthermore,  We conduct a human evaluation survey on whether extracted stages are physically valid. Each of the 50 videos was rated by all 5 raters, giving 250 individual ratings, and the resulting estimates are precise with mean pairwise percent agreement of $0.853$ , with inter-rater agreement of $0.861$ using Gwet's AC2.

\begin{table}[t]
\centering
\small
\begin{tabular}{l c c c c}
\toprule
\textbf{Physics Domain} & \textbf{\#Videos} & \textbf{Stage Quality} & \textbf{Ordering} & \textbf{Overall} \\
 & & \textit{mean $\pm$ 95\% CI} & \textit{mean $\pm$ 95\% CI} & \textit{mean} \\
\midrule
Rigid-body / collision  & 11 & 4.42 $\pm$ 0.21 & 4.61 $\pm$ 0.23 & 4.46\\
Fluid dynamics          & 12 & 4.31 $\pm$ 0.23 & 4.55 $\pm$ 0.25 & 4.36\\
Thermodynamics          & 8  & 4.05 $\pm$ 0.30 & 4.28 $\pm$ 0.32 & 4.10\\
Deformation             & 5  & 4.24 $\pm$ 0.34 & 4.40 $\pm$ 0.36 & 4.27\\
Optics                  & 8  & 3.88 $\pm$ 0.32 & 4.12 $\pm$ 0.34 & 3.92\\
Combustion              & 6  & 4.18 $\pm$ 0.31 & 4.45 $\pm$ 0.33 & 4.24\\
\midrule
\textbf{Overall}        & \textbf{50} & \textbf{4.21 $\pm$ 0.10} & \textbf{4.42 $\pm$ 0.11} & \textbf{4.25} \\
\bottomrule
\end{tabular}

\caption{Human validation of extracted sub-stages across $n=50$ videos and 5 raters. Each video was rated by all 5 raters on a 1–5 Likert scale. We report mean ratings with 95\% confidence intervals for individual sub-stage rendering quality (averaged across all sub-stages of a video) and temporal ordering.}
\label{tab:human_eval_stage}
\end{table}

Additionally, to validate physical structure, 10 physics graduate raters evaluated decompositions for 20 canonical events, with 10 corrupted trees interspersed and unmarked; 5 with stages shuffled out of chronological order, 5 with a physically impossible stage inserted. Raters were told only that some decompositions contained errors. Table \ref{tab:table_f} shows the average score collected from these 10 raters. As seen, the extracted trees score comparably and the corrupted variants scores are low, with the largest gap on ordering, where shuffling the stages is the most direct violation of physical sequence. This confirms that raters were evaluating physical validity, and that our reference decompositions reflect the physical structure of each event.
\begin{table}[htbp]
\centering
\small
\setlength{\tabcolsep}{8pt}
\begin{tabular}{lccc}
\toprule
 & \textbf{Physical correctness} & \textbf{Ordering} & \textbf{Completeness} \\
\midrule
Extracted tree & \textbf{4.12} & \textbf{3.78} & \textbf{3.86} \\
Corrupted tree & 2.23          & 1.80          & 2.40          \\
\bottomrule
\end{tabular}
\caption{Human evaluation results for extracted versus corrupted stage trees.}
\label{tab:table_f}
\end{table}

\section{Hierarchical vs. flat rewards.} 
We remove the sub-stages from each reference tree, keeping only the top-level stages. This gives a flat reward and is the depth-one special case of our formulation. The flat reward barely hurts 
VideoPhy2, where each prompt has one physical principle, but degrades MultiPhyBench sharply. The reason is that a top-level stage such as "splash" is satisfied by any splash at all. Without sub-stages the reward cannot separate a static spray from a real impact followed by crown formation, so it stops pushing the model once each principle is roughly present. This matters most when several principles must unfold at once.
\begin{table}[t]
\centering
\small
\setlength{\tabcolsep}{4.4pt}
\begin{tabular}{lccc}
\toprule
Benchmark & Variant & PC & SA \\
\midrule
VideoPhy2 & Hierarchical & 66.0 & 52.0 \\
VideoPhy2 & Flat & 63.0 & 47.0 \\
MultiPhyBench & Hierarchical & 59.0 & 44.0 \\
MultiPhyBench & Flat & 47.0 & 29.0 \\
\bottomrule
\end{tabular}
\caption{Ablation on reward structure}
\label{tab:reward_ablation}
\end{table}

\section{Ablation on individual alignment, completeness, and ordering}
We ablate each term of the hierarchical reward (Eq. 5) and evaluate on MultiPhyBench using three diagnostics, each targeting one term. Omission rate measures the fraction of physical principles absent from the video, ordering accuracy is the fraction of matched sub-stages in correct relative sequence, and stage match is the mean similarity between reference and observed stages. Reported scores are over 30 videos per variant, evaluated by Gemini Flash. To confirm these automatic diagnostics are reliable, five human raters independently scored the videos drawn evenly across variants; agreement between the human and automatic scores are also reported at the bottom of Table \ref{tab:table_b}. Removing a term degrades its own diagnostic far more than the others, confirming \begin{table}[htbp]
\centering
\small
\setlength{\tabcolsep}{6pt}
\begin{tabular}{lccc}
\toprule
\textbf{Reward Variant} & \textbf{Omission rate (\%)} $\downarrow$ & \textbf{Ordering accuracy (\%)} $\uparrow$ & \textbf{Stage match (\%)} $\uparrow$ \\
\midrule
Full             & \textbf{6.1} & \textbf{86.0} & \textbf{81.0} \\
\midrule
($-$) ordering     & 6.8          & 61.0          & 72.0          \\
($-$) completeness & 17.4         & 84.0          & 79.0          \\
($-$) alignment    & 9.7          & 83.0          & 68.0          \\
\bottomrule
\end{tabular}
\caption{Gemini--human agreement is computed as the fraction of items on which the automatic and human labels match: 0.84 (omission), 0.77 (ordering), 0.82 (stage match).}
\label{tab:table_b}
\end{table}that the three terms supervise distinct properties. Two smaller cross-effects are expected. 
Out-of-order processes produce less coherent extracted stage descriptions, which lowers stage match, and alignment is normalized by the total reference leaf count, so it carries a soft completeness signal.

\section{Ablation on Normalization vs. scaling}
The normalization and scaling serve two distinct purposes in our formulation, isolated in Table \ref{tab:table_c} below. Per-stream normalization controls scale and variance differences among reward streams across the rollouts for a prompt. \begin{table}[htbp]
\centering
\small
\setlength{\tabcolsep}{6pt}
\begin{tabular}{lcccc}
\toprule
\textbf{Exp3} & \textbf{PC (N=1)} & \textbf{PC (N=2)} & \textbf{PC (N=3)} & \textbf{Overall SA} \\
\midrule
Full                        & 72.1 & \textbf{63.0} & \textbf{48.0} & \textbf{58.3} \\
w/o per-stream norm         & 70.3 & 38.0          & 22.3          & 21.6          \\
w/o $1/\sqrt{n}$ rescaling  & 72.8 & 41.3          & 37.6          & 35.0          \\
\bottomrule
\end{tabular}
\caption{Ablation results on normalization and rescaling components.}
\label{tab:table_c}
\end{table}
Each stream is standardized against its own mean and standard deviation over the group, so every principle contributes comparably to the advantage. Without it, a stream that varies more widely dominates and the policy maximizes that one term while ignoring the rest, degrading alignment and multi-process scenes (second row). The $\frac{1}{\sqrt{n}}$ rescaling addresses a different problem. Summing more terms inflates the variance of the advantage, so a three-principle sample pushes the gradient harder than a one-principle sample; the rescaling holds this variance constant. It also preserves the balance between the two reward families, since the advantage combines one global stream with a sum of n local ones. Without the factor the local sum overwhelms the global term and the policy loses the signal anchoring generation to prompt semantics and visual coherence, visible as the SA drop in the third row. At $N = 1$ the factor is inactive and per-stream normalization has little to balance, with only one local stream alongside the global one; the differences emerge as principles co-occur.

\section{User Study}

\begin{figure*}[!htb]
\centering
\includegraphics[width=0.85\textwidth]{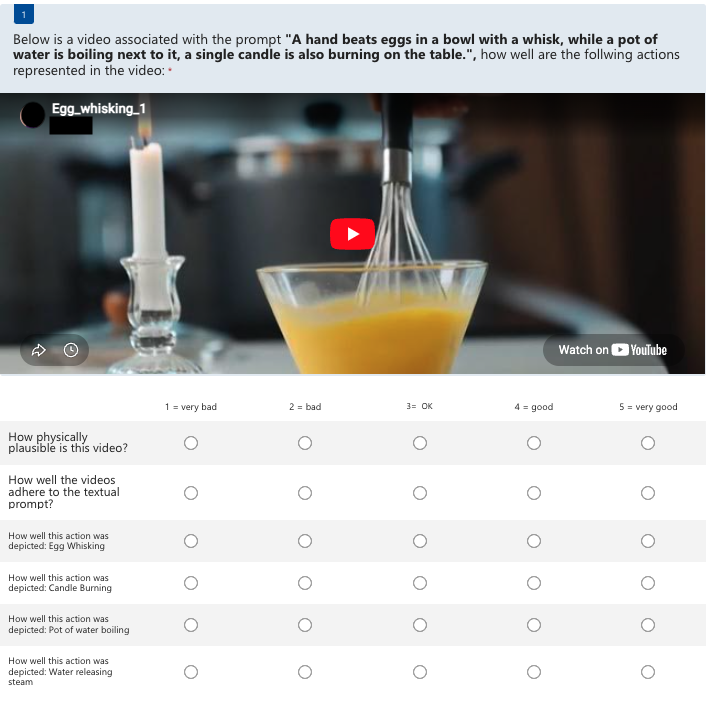}
    \vspace{-1em}
    \caption{\textbf{User study sample question}}
    \vspace{-1em}
\label{fig:user_study_question}
\end{figure*}

To complement the automatic evaluations, we analyze survey results for videos containing multiple concurrent physical principles. We report the average ratings from 50 participants who evaluated the depiction of each physical event; a representative question from this study is shown in Fig.~\ref{fig:user_study_question}. The user study included 30 distinct prompts across 5 different methods. As illustrated in Fig.~\ref{fig:per-process}, A--C, our method consistently achieves the highest scores across all principles. Notably, while baseline performance is often unstable, dropping specific principles entirely within a scene or the inability to generate all of them plausibly, our approach maintains balanced adherence across the board. This validates our per-process reward design, which successfully allocates a distinct learning signal to each physical event.

 \section{Error Bars for the quantitative comparisons}
 Table \ref{tab:combined-evaluation} includes all the standard deviations as the error bar for the main quantitative comparisons and ablations. To enhance readability, we present the error bars in the supplementary rather than the main table.

 \begin{table*}[t]
\centering
\caption{Comprehensive evaluation across VideoPhy2, PhyWorldBench, Multi-concept, and VBench along with their standard deviations.}
\label{tab:combined-evaluation}
\resizebox{\textwidth}{!}{
\setlength{\tabcolsep}{2pt}
\begin{tabular}{l|ccc|ccc|ccc|cc}
\toprule
 & \multicolumn{3}{c|}{\textbf{VideoPhy2}} & \multicolumn{3}{c|}{\textbf{PhyWorldBench}} & \multicolumn{3}{c|}{\textbf{Multi-concept}} & \multicolumn{2}{c}{\textbf{VBench}} \\
 \cmidrule(lr){2-4} \cmidrule(lr){5-7} \cmidrule(lr){8-10} \cmidrule(l){11-12}
Method & PC$\uparrow$ & SA$\uparrow$ & Both$\uparrow$ & PC$\uparrow$ & SA$\uparrow$ & Both$\uparrow$ & PC$\uparrow$ & SA$\uparrow$ & Both$\uparrow$ & Dyn.$\uparrow$ & App.$\uparrow$ \\
\midrule
Wan2.1    & 49.10\tiny{$\pm$2.06} & 43.60\tiny{$\pm$2.04} & 30.50\tiny{$\pm$1.90} & 47.7\tiny{$\pm$2.91} & 40.9\tiny{$\pm$2.86} & 23.7\tiny{$\pm$2.48} & 53.0\tiny{$\pm$2.88} & 35.0\tiny{$\pm$2.75} & 27.0\tiny{$\pm$2.56} & 0.808\tiny{$\pm$0.014} & \underline{0.573\tiny{$\pm$0.003}}\\
PnP       & \underline{80.1\tiny{$\pm$1.64}} & \underline{68.0\tiny{$\pm$1.92}} & \underline{60.14\tiny{$\pm$2.02}} & 56.1\tiny{$\pm$2.89} & 48.5\tiny{$\pm$2.91} & 31.2\tiny{$\pm$2.70} & \underline{57.0\tiny{$\pm$2.86}} & \underline{37.0\tiny{$\pm$2.79}} & \underline{29.0\tiny{$\pm$2.62}} & 0.844\tiny{$\pm$0.013} & 0.549\tiny{$\pm$0.003}\\
VideoRepa & 70.9\tiny{$\pm$1.87} & 53.5\tiny{$\pm$2.05} & 43.8\tiny{$\pm$2.04} & 51.7\tiny{$\pm$2.91} & 33.3\tiny{$\pm$2.74} & 19.3\tiny{$\pm$2.30} & 49.0\tiny{$\pm$2.89} & 22.0\tiny{$\pm$2.39} & 15.0\tiny{$\pm$2.06} & 0.863\tiny{$\pm$0.012} & 0.550\tiny{$\pm$0.003}\\
PhyT2V    & 79.0\tiny{$\pm$1.68} & 65.0\tiny{$\pm$1.96} & 55.0\tiny{$\pm$2.05} & 40.3\tiny{$\pm$2.86} & 47.5\tiny{$\pm$2.91} & 22.7\tiny{$\pm$2.44} & 50.0\tiny{$\pm$2.89} & 26.0\tiny{$\pm$2.53} & 15.0\tiny{$\pm$2.06} & 0.872\tiny{$\pm$0.012} & 0.500\tiny{$\pm$0.003}\\
WISA      & 68.0\tiny{$\pm$1.92} & 59.7\tiny{$\pm$2.02} & 44.3\tiny{$\pm$2.05} & 56.3\tiny{$\pm$2.89} & 50.7\tiny{$\pm$2.91} & 30.7\tiny{$\pm$2.69} & 52.5\tiny{$\pm$2.88} & 43.0\tiny{$\pm$2.86} & 24.5\tiny{$\pm$2.48} & 0.859\tiny{$\pm$0.012} & 0.567\tiny{$\pm$0.003}\\
GPT-5     & 69.2\tiny{$\pm$1.90} & 60.9\tiny{$\pm$2.01} & 43.1\tiny{$\pm$2.04} & \underline{56.9\tiny{$\pm$2.88}} & \underline{52.5\tiny{$\pm$2.91}} & \underline{31.8\tiny{$\pm$2.71}} & 51.5\tiny{$\pm$2.89} & 44.0\tiny{$\pm$2.87} & 24.5\tiny{$\pm$2.48} & \underline{0.882\tiny{$\pm$0.011}} & 0.531\tiny{$\pm$0.003}\\
\rowcolor[HTML]{F0F5FF} Ours (T2V) & \textbf{88.6\tiny{$\pm$1.31}} & \textbf{79.2\tiny{$\pm$1.67}} & \textbf{68.8\tiny{$\pm$1.91}} & \textbf{74.9\tiny{$\pm$2.52}} & \textbf{76.3\tiny{$\pm$2.48}} & \textbf{61.6\tiny{$\pm$2.83}} & \textbf{76.7\tiny{$\pm$2.44}} & \textbf{68.85\tiny{$\pm$2.67}} & \textbf{52.7\tiny{$\pm$2.88}} & \textbf{0.905\tiny{$\pm$0.010}} & \textbf{0.592\tiny{$\pm$0.003}}\\
\midrule
Qwen 2.5  & 60.4\tiny{$\pm$2.01} & 62.9\tiny{$\pm$1.99} & 41.5\tiny{$\pm$2.03} & 58.6\tiny{$\pm$2.87} & 58.3\tiny{$\pm$2.87} & 44.5\tiny{$\pm$2.89} & 50.3\tiny{$\pm$2.89} & 39.2\tiny{$\pm$2.82} & 19.1\tiny{$\pm$2.27} & 0.874\tiny{$\pm$0.012} & 0.517\tiny{$\pm$0.003} \\
Ours (global) & \underline{72.7\tiny{$\pm$1.83}} & \textbf{75.3\tiny{$\pm$1.78}} & \underline{56.3\tiny{$\pm$2.04}} & \underline{62.3\tiny{$\pm$2.82}} & \textbf{64.3\tiny{$\pm$2.79}} & \underline{54.3\tiny{$\pm$2.90}} & 61.0\tiny{$\pm$2.82} & 53.0\tiny{$\pm$2.88} & 40.0\tiny{$\pm$2.83} & 0.871\tiny{$\pm$0.012} & \textbf{0.576\tiny{$\pm$0.003}}\\
Ours (local, sum) & 72.2\tiny{$\pm$1.84} & 70.1\tiny{$\pm$1.88} & 48.5\tiny{$\pm$2.06} & 61.9\tiny{$\pm$2.83} & 57.3\tiny{$\pm$2.88} & 37.1\tiny{$\pm$2.81} & \underline{67.0\tiny{$\pm$2.71}} & \underline{57.9\tiny{$\pm$2.85}} & \underline{42.4\tiny{$\pm$2.85}} & 0.865\tiny{$\pm$0.012} & 0.556\tiny{$\pm$0.003} \\
Ours (local, normalized) & \textbf{79.2\tiny{$\pm$1.67}} & \underline{70.4\tiny{$\pm$1.88}} & \textbf{57.3\tiny{$\pm$2.04}} & \textbf{65.6\tiny{$\pm$2.77}} & \underline{60.8\tiny{$\pm$2.84}} & \textbf{52.5\tiny{$\pm$2.91}} & \textbf{68.5\tiny{$\pm$2.68}} & \textbf{58.3\tiny{$\pm$2.85}} & \textbf{44.5\tiny{$\pm$2.87}} & \textbf{0.893\tiny{$\pm$0.011}} & 0.567\tiny{$\pm$0.003} \\
\bottomrule
\end{tabular}
}
\end{table*}
\clearpage
\section{Stage Generation Prompt}
\label{sec:stage_prompt}

We use the following prompt to instruct an LLM to decompose each video description into distinct physical processes and per-process temporal stages. The output is parsed as JSON and used as the supervision target $T^{*(i)}$ for our hierarchical training (Sec.~\ref{sec:reward}).

\begin{lstlisting}[caption={Stage generation prompt used to produce reference stage trees $T^{*(i)}$.}, label={lst:stage_prompt}]
{Stage Generation Prompt}
You are a physics expert analyzing a video. Given the description and physics context below, decompose the video into distinct physical processes, and break down each process into its own ordered stages.

**Video Description:**
{prompt}

**Category:** {category}



**Instructions:**
1. Identify all distinct physical processes or interacting entities occurring in the video (e.g., "steel_marble_kinematics" and "viscous_honey_displacement").
2. For EACH process, decompose its temporal progression into chronological stages.
3. Stages within a process must represent physically meaningful transitions (e.g., initial state, momentum transfer, deformation, settling).
4. Provide 2-5 stages per process, depending on its complexity.
5. Each stage name should be a concise identifier (2-4 words, snake_case).
6. The description must specify the observable forces, energy transfers, or material interactions unique to that specific process.

**Respond in this exact JSON format and nothing else:**

{
    "processes": [
        {
            "process_name": "identifier_for_process_1",
            "physics_domain": "rigid_body | fluid_dynamics | thermodynamics | deformation | optics | acoustics | biological | electromagnetic | multi_domain",
            "stages": [
                {"stage": 1, "name": "stage_name", "description": "What physically happens and what is visually observable for this entity"},
                {"stage": 2, "name": "stage_name", "description": "..."}
            ]
        },
        {
            "process_name": "identifier_for_process_2",
            "physics_domain": "...",
            "stages": [
                {"stage": 1, "name": "...", "description": "..."},
                {"stage": 2, "name": "...", "description": "..."}
            ]
        }
    ]
}
\end{lstlisting}
\clearpage
\section{Video Prompts}
Tables \ref{tab:sponge} to \ref{tab:coin} present the sample prompts generated by \texttt{HiPhy} that are associated with the visuals in the main paper.
\begin{table*}[t]
\centering
\caption{Video Prompts}
\label{tab:sponge}

\begin{tcolorbox}[
    title=\textbf{A sponge is squeezed over a bowl of water while bubbles rise in a fish tank beside it.}, 
    colframe=blue!60!black, 
    colback=white,           
    coltitle=white,          
    fonttitle=\bfseries, 
    width=\textwidth,        
    arc=2mm,                 
    boxrule=1.5pt            
]
\ttfamily \scriptsize 
\hspace*{4mm} \textbf{HiPhy Expanded}: A saturated sponge is gripped and compressed between both hands, the applied pressure collapsing the open pore structure and forcing the water stored within out through the surface in multiple thin streams that converge and fall into the bowl below, the sponge body deforming visibly under the hand pressure and springing partially back as the grip relaxes and air rushes back into the reopening pores. Beside it in a fish tank an air stone releases a continuous stream of small air bubbles from the tank base, each bubble rising through the water column and expanding slightly as the water pressure decreases with decreasing depth, the bubbles reaching the surface and bursting to release small ripples that spread across the tank water surface.

\end{tcolorbox}
\end{table*}

\begin{table*}[t]
\centering
\caption{Video Prompts}
\label{tab:hammer}

\begin{tcolorbox}[
    title=\textbf{A hammer strikes a nail into wood while a pot of water boils on the stove beside it.}, 
    colframe=blue!60!black, 
    colback=white,           
    coltitle=white,          
    fonttitle=\bfseries, 
    width=\textwidth,        
    arc=2mm,                 
    boxrule=1.5pt            
]
\ttfamily \scriptsize 
\hspace*{4mm} \textbf{HiPhy Expanded}: A hammer is swung downward and its steel head makes hard contact with the nail head, driving the nail shaft into the wood grain with each strike as the wood fibers compress and split around the penetrating metal, the nail advancing incrementally with each blow while the hammer rebounds slightly from the elastic response of the steel on steel contact. Simultaneously on the stove beside the workbench a pot of water boils vigorously, with dense clusters of bubbles nucleating at the heated base and bursting at the surface releasing continuous steam, the vibration transmitted through the floor from each hammer blow causing the boiling water surface to ripple and the steam column above to shudder briefly with each impact.

\end{tcolorbox}
\end{table*}

\begin{table*}[t]
\centering
\caption{Video Prompts}
\label{tab:balloon}

\begin{tcolorbox}[
    title=\textbf{A balloon floats upward while steam rises from a pot on the stove below it.}, 
    colframe=blue!60!black, 
    colback=white,           
    coltitle=white,          
    fonttitle=\bfseries, 
    width=\textwidth,        
    arc=2mm,                 
    boxrule=1.5pt            
]
\ttfamily \scriptsize 
\hspace*{4mm} \textbf{HiPhy Expanded}: A helium-filled balloon floats upward toward the ceiling, the buoyancy force from the less dense helium inside exceeding the downward gravitational pull on the balloon's total mass, the balloon drifting upward and rotating slowly as minor air currents in the room push it laterally. Below it on the stove a pot of boiling water releases a continuous column of steam upward from the surface, the hot water vapor rising rapidly due to its lower density than the surrounding cooler air, the rising steam column reaching the underside of the floating balloon and deflecting outward around its curved surface, slightly accelerating the balloon's upward drift as the warm rising air adds to the buoyancy force beneath it.

\end{tcolorbox}
\end{table*}

\begin{table*}[t]
\centering
\caption{Video Prompts}
\label{tab:rubber_ball}

\begin{tcolorbox}[
    title=\textbf{A colorful rubber ball is dropped from a height, showing the bounce of the ball as it makes contact with the hard floor.}, 
    colframe=blue!60!black, 
    colback=white,           
    coltitle=white,          
    fonttitle=\bfseries, 
    width=\textwidth,        
    arc=2mm,                 
    boxrule=1.5pt            
]
\ttfamily \scriptsize 
\hspace*{4mm} \textbf{HiPhy Expanded}: the video captures view of one colorful rubber ball being dropped from a height. the ball is colorful, creating a vibrant and eye-catching display. as the ball descends, it bounces off the hard floor, causing to bounce lightly. the video ends with the ball lying flat on the ground, demonstrating the energy and impact of the fall.

\end{tcolorbox}
\end{table*}

\begin{table*}[t]
\centering
\caption{Video Prompts}
\label{tab:apple}

\begin{tcolorbox}[
    title=\textbf{Knife skillfully slices an apple.}, 
    colframe=blue!60!black, 
    colback=white,           
    coltitle=white,          
    fonttitle=\bfseries, 
    width=\textwidth,        
    arc=2mm,                 
    boxrule=1.5pt            
]
\ttfamily \scriptsize 
\hspace*{4mm} \textbf{HiPhy Expanded}: the knife skillfully slices through the juicy apple, which is positioned upright on a wooden cutting board. the apple is bright red and glistening with moisture, indicating its freshness. as the knife cuts through the apple, it lifts off the cutting board, revealing the depth of the cut. the overall scene emphasizes the artistry involved in slicing through a ripe apple with precision.

\end{tcolorbox}
\end{table*}

\begin{table*}[t]
\centering
\caption{Video Prompts}
\label{tab:leaf}

\begin{tcolorbox}[
    title=\textbf{A leaf falls from a tree while rain hits a puddle below.}, 
    colframe=blue!60!black, 
    colback=white,           
    coltitle=white,          
    fonttitle=\bfseries, 
    width=\textwidth,        
    arc=2mm,                 
    boxrule=1.5pt            
]
\ttfamily \scriptsize 
\hspace*{4mm} \textbf{HiPhy Expanded}: A dry leaf detaches from a branch and falls under gravity, its flat irregular shape and low mass making it susceptible to air resistance and causing it to tilt, rotate, and drift laterally in a slow oscillating descent rather than falling straight, the leaf tumbling end over end and drifting sideways as pockets of air deflect its trajectory unpredictably. Below it rain strikes a puddle on the ground at high velocity, each raindrop impact generating a small crown splash with thin water jets ejecting upward before collapsing, and sending circular ripple rings radiating outward across the puddle surface, the falling leaf eventually landing on the puddle surface where surface tension briefly supports it before it becomes waterlogged and sinks flush with the water surface.

\end{tcolorbox}
\end{table*}

\begin{table*}[t]
\centering
\caption{Video Prompts}
\label{tab:baseball}

\begin{tcolorbox}[
    title=\textbf{A glove catching the fast-moving baseball.}, 
    colframe=blue!60!black, 
    colback=white,           
    coltitle=white,          
    fonttitle=\bfseries, 
    width=\textwidth,        
    arc=2mm,                 
    boxrule=1.5pt            
]
\ttfamily \scriptsize 
\hspace*{4mm} \textbf{HiPhy Expanded}: the video captures a close-up view of a glove catching a fast-moving baseball. the glove is positioned on the left hand side of the frame, while the baseball travels from right to left. as the baseball approaches, the glove extends its fingers to grip the ball firmly. the glove's movement is smooth and deliberate, ensuring a secure catch. the background is blurred, revealing a playing field with grass and possibly other sports equipment visible in the distance. the lighting suggests it might be daytime, adding to the dynamic atmosphere of the scene.

\end{tcolorbox}
\end{table*}
\begin{table*}[t]
\centering
\caption{Video Prompts}
\label{tab:gymnast}

\begin{tcolorbox}[
    title=\textbf{Gymnast performs an aerial somersault at sunrise.}, 
    colframe=blue!60!black, 
    colback=white,           
    coltitle=white,          
    fonttitle=\bfseries, 
    width=\textwidth,        
    arc=2mm,                 
    boxrule=1.5pt            
]
\ttfamily \scriptsize 
\hspace*{4mm} \textbf{HiPhy Expanded}: A gymnast sprints forward and plants both feet simultaneously into the ground, driving both knees upward and swinging both arms overhead to convert forward momentum into vertical lift, launching the body upward into the air with no hand contact with the ground. At peak height the gymnast pulls both knees tightly into the chest and wraps both arms around the shins, tucking the body into a compact rotating mass that completes one full forward rotation around a horizontal axis through the hips. As the body descends and the rotation completes, the legs extend downward and outward to prepare for landing, both feet striking the ground simultaneously with knees bent deeply to absorb the downward impact force, the body straightening upright to hold the landing position against the warm early morning sky.

\end{tcolorbox}
\end{table*}

\begin{table*}[t]
\centering
\caption{Video Prompts}
\label{tab:skateboard}

\begin{tcolorbox}[
    title=\textbf{A skateboarder performs jumps while splashing through a large puddle on a street.}, 
    colframe=blue!60!black, 
    colback=white,           
    coltitle=white,          
    fonttitle=\bfseries, 
    width=\textwidth,        
    arc=2mm,                 
    boxrule=1.5pt            
]
\ttfamily \scriptsize 
\hspace*{4mm} \textbf{HiPhy Expanded}: the video captures a dramatic scene of a skateboard splashing through a large puddle on a street. the skateboarder is dressed in a black shirt, blue jeans, and white sneakers. the skateboard, featuring a yellow and black striped deck, is positioned on the left side of the frame. as the video progresses, the skateboarder begins to accelerate, causing the skateboard to splash vigorously through the puddle. the splashes create a dramatic contrast against the dark background of the street. the skateboarder's movement adds to the sense of chaos and urgency in the scene. the video concludes with the skateboarder continuing to ride through the puddle, creating a continuous stream of water as it flows downstream. the background remains dark, emphasizing the vibrant colors and movement of the water and the skateboarder. there are no animals or human characters present in the video, focusing solely on the interaction between the skateboarder, the water, and the urban setting. the video captures the raw energy and fun of skateboarding in a dynamic outdoor environment.

\end{tcolorbox}
\end{table*}

\begin{table*}[t]
\centering
\caption{Video Prompts}
\label{tab:honey}

\begin{tcolorbox}[
    title=\textbf{A stick of incense burns while honey drips slowly from a spoon beside it.}, 
    colframe=blue!60!black, 
    colback=white,           
    coltitle=white,          
    fonttitle=\bfseries, 
    width=\textwidth,        
    arc=2mm,                 
    boxrule=1.5pt            
]
\ttfamily \scriptsize 
\hspace*{4mm} \textbf{HiPhy Expanded}: A stick of incense burns at its tip with a small glowing ember, releasing a thin continuous thread of white smoke that rises straight upward in a slow coherent column before breaking into irregular curling wisps as it reaches the turbulent ambient air above, the smoke dispersing and spreading outward as it cools and loses buoyancy. Beside it honey drips from a spoon in a slow thick strand that stretches and thins under gravity, the high viscosity of the honey resisting the flow and causing the strand to elongate into a long thin thread before detaching and falling, the detached drop coiling slowly as it lands on the surface below while the rising smoke from the incense bends slightly toward the warmth of the honey.

\end{tcolorbox}
\end{table*}

\begin{table*}[t]
\centering
\caption{Video Prompts}
\label{tab:toothpaste}

\begin{tcolorbox}[
    title=\textbf{Camera focuses on a toothpaste tube while a hand squeezes a steady stream of toothpaste.}, 
    colframe=blue!60!black, 
    colback=white,           
    coltitle=white,          
    fonttitle=\bfseries, 
    width=\textwidth,        
    arc=2mm,                 
    boxrule=1.5pt            
]
\ttfamily \scriptsize 
\hspace*{4mm} \textbf{HiPhy Expanded}: A finger presses down on the flexible body of a toothpaste tube, deforming the plastic casing inward and forcing the viscous paste toward the nozzle where it resists briefly before yielding and extruding in a smooth continuous ribbon that lands and coils onto the bristles of a toothbrush held below. The ribbon conforms slightly to the gaps between the bristle tufts under its own weight without spreading or dripping, and when finger pressure is released the flow stops abruptly with the paste holding its coiled shape on the brush due to its yield stress preventing further movement.

\end{tcolorbox}
\end{table*}
\begin{table*}[t]
\centering
\caption{Video Prompts}
\label{tab:book}

\begin{tcolorbox}[
    title=\textbf{A book slides off a tilted shelf and falls to the ground while a small rubber ball bounces repeatedly on the desk.}, 
    colframe=blue!60!black, 
    colback=white,           
    coltitle=white,          
    fonttitle=\bfseries, 
    width=\textwidth,        
    arc=2mm,                 
    boxrule=1.5pt            
]
\ttfamily \scriptsize 
\hspace*{4mm} \textbf{HiPhy Expanded}: the video captures a dynamic scene set on a desk. the focal point is a book that slides off a tilted shelf. as the book falls, it creates a dramatic contrast against the plain white background of the desk. in the background, a small rubber ball bounces on the desk, adding a playful element to the otherwise focused scene. the overall atmosphere is one of tension and surprise, with the sudden fall of the book creating a stark contrast against the static background.

\end{tcolorbox}
\end{table*}

\begin{table*}[t]
\centering
\caption{Video Prompts}
\label{tab:coffee}

\begin{tcolorbox}[
    title=\textbf{Coffee accepting a gentle pour of milk.}, 
    colframe=blue!60!black, 
    colback=white,           
    coltitle=white,          
    fonttitle=\bfseries, 
    width=\textwidth,        
    arc=2mm,                 
    boxrule=1.5pt            
]
\ttfamily \scriptsize 
\hspace*{4mm} \textbf{HiPhy Expanded}: the video depicts a close-up view of a coffee mug. inside the mug, there is a small amount of coffee. the mug is being gently poured into by a hand holding a clear glass bottle. the liquid being poured appears to be milk or a similar liquid. as the liquid is poured, it creates a gentle frothy mixture inside the mug. the overall scene remains static throughout the video, focusing on the interaction between the liquid being poured and the existing contents of the mug.

\end{tcolorbox}
\end{table*}

\begin{table*}[t]
\centering
\caption{Video Prompts}
\label{tab:sauce_simmers}

\begin{tcolorbox}[
    title=\textbf{Sauce simmers in a pan on the stove releasing steam, at the ironing board an iron presses down on a shirt, and through the kitchen window wind drives rain against the glass.}, 
    colframe=blue!60!black, 
    colback=white,           
    coltitle=white,          
    fonttitle=\bfseries, 
    width=\textwidth,        
    arc=2mm,                 
    boxrule=1.5pt            
]
\ttfamily \scriptsize 
\hspace*{4mm} \textbf{HiPhy Expanded}: A cabin interior frames two adjacent tables and a rain streaked window. on the left, a saucepan of tomato sauce undergoes low intensity convective boiling on an active burner, with small bubbles nucleating at the base and bursting at the viscous surface, releasing intermittent steam plumes that rise and disperse into the warm air. On the right table, a cotton shirt is pressed under a heated iron driven in firm, repetitive linear strokes. through the back window, dense near vertical rainfall streaks downward under gravity, producing crown splashes and ripple rings on pooling backyard surfaces while irregular rivulets distort the gloomy view through the glass.

\end{tcolorbox}
\end{table*}

\begin{table*}[t]
\centering
\caption{Video Prompts}
\label{tab:egg_candle}

\begin{tcolorbox}[
    title=\textbf{A hand beats egg in a bowl with a whisk, while a pot of water is boiling next to it, a single candle is also burning on the table}, 
    colframe=blue!60!black, 
    colback=white,           
    coltitle=white,          
    fonttitle=\bfseries, 
    width=\textwidth,        
    arc=2mm,                 
    boxrule=1.5pt            
]
\ttfamily \scriptsize 
\hspace*{4mm} \textbf{HiPhy Expanded}: On a kitchen counter top, A hand beats egg white in a bowl with a whisk, the liquid thickening into white foam with each stroke. On the stove beside it, a pot of water is boiling with bubbles breaking at the surface and releasing steady puffs of steam. On the counter top next to the stove, a single candle burns on the table with a small steady flame.

\end{tcolorbox}
\end{table*}

\begin{table*}[t]
\centering
\caption{Video Prompts}
\label{tab:coin}

\begin{tcolorbox}[
    title=\textbf{A coin dances in a spiral on the table}, 
    colframe=blue!60!black, 
    colback=white,           
    coltitle=white,          
    fonttitle=\bfseries, 
    width=\textwidth,        
    arc=2mm,                 
    boxrule=1.5pt            
]
\ttfamily \scriptsize 
\hspace*{4mm} \textbf{HiPhy Expanded}: A coin dances in a spiral on the table.": "the video captures a close-up view of a coin dancing in a spiral on the table. the coin is silver and features a portrait of queen elizabeth II on its obverse side. the background shows a plain white surface, likely a table or countertop. as the video progresses, the coin continues to dance in a spiral pattern. the spiral is tight at the beginning and gradually expands as the coin moves further away from its initial position. the movement of the coin is smooth and continuous, creating a visually appealing and mesmerizing dance.

\end{tcolorbox}
\end{table*}

\end{document}